%% file: neurips_2024.tex
\UseRawInputEncoding
\documentclass{article}

     \PassOptionsToPackage{numbers, compress}{natbib}

\usepackage[preprint]{neurips_2024}

\usepackage[utf8]{inputenc} 
\usepackage[T1]{fontenc}
\usepackage[colorlinks=true]{hyperref}
\usepackage{url}
\usepackage{booktabs}       
\usepackage{amsfonts}       
\usepackage{lipsum} 
\usepackage{xspace}
\usepackage{amsmath}        
\usepackage{nicefrac}
\usepackage{microtype}
\usepackage{xcolor}
\usepackage{graphicx}       
\usepackage{wrapfig}
\usepackage{siunitx} 
\usepackage{caption}     

\usepackage{tcolorbox}

\usepackage{fancyvrb}
\usepackage{subcaption}

\newcommand{\meanstd}[2]{#1 \!{\scriptsize $\pm$\! #2}}

\definecolor{citecolor}{HTML}{3498DB}
\definecolor{urlcolor}{HTML}{485DFF}

\tcbuselibrary{listings}
\tcbuselibrary{breakable}

\lstdefinestyle{promptstyle}{
  basicstyle=\footnotesize\fontfamily{zi4}\selectfont,
  columns=fullflexible,
  escapechar=§,
  breaklines=true,
  breakautoindent=false,  
  breakindent=0pt,        
  showstringspaces=false,
  tabsize=2
}

\definecolor{codegreen}{rgb}{0,0.6,0}
\definecolor{codegray}{rgb}{0.5,0.5,0.5}
\definecolor{codepink}{RGB}{252, 142, 172}
\definecolor{codepurple}{rgb}{0.58,0,0.82}
\definecolor{backcolour}{RGB}{245,245,245}
\lstdefinestyle{mystyle}{
    backgroundcolor=\color{backcolour},   
    commentstyle=\color{magenta},
    keywordstyle=\color{blue},
    numberstyle=\tiny\color{codegray},
    stringstyle=\color{codepurple},
    basicstyle=\fontfamily{\ttdefault}\footnotesize,
    breakatwhitespace=false,         
    breaklines=true,                 
    literate={³}{$^3$}1
         {²}{$^2$}1
         {—}{---}1,
    keepspaces=true,    
    frame=single,
    numbersep=5pt,                  
    showspaces=false,                
    showstringspaces=false,
    showtabs=false,                  
    tabsize=2,
    classoffset=1, %
    keywordstyle=\color{violet},
    classoffset=0,
}

\usepackage{placeins}

\hypersetup{citecolor=citecolor,
urlcolor=urlcolor}

\usepackage{enumitem}
\setitemize{noitemsep,topsep=0pt,parsep=0pt,partopsep=0pt, leftmargin=*}
\setenumerate{noitemsep,topsep=0pt,parsep=0pt,partopsep=0pt, leftmargin=*}

\newcommand{\numclasses}{10\xspace}        
\newcommand{\nummaterials}{5\xspace}        
\newcommand{\numobjects}{\num{1624}\xspace}  

\definecolor{correctgreen}{HTML}{228B22}  
\definecolor{incorrectred}{HTML}{D62728}  

\newif\ifrebuttal
\rebuttaltrue            

\definecolor{rebuttalcol}{HTML}{000000} 

\let\rebutal\rebuttal    

\newcommand{\method}{\textsc{Pixie}\xspace}
\newcommand{\dataset}{\textsc{PixieVerse}\xspace}
\title{Pixie: Fast and Generalizable Supervised Learning of 3D Physics from Pixels}
\newcommand{\weburl}{\url{https://pixie-3d.github.io/}}

\newcommand{\gain}{1.46-4.39x }
\author{%
  Long Le$^{1}$\thanks{Correspondence: \texttt{vlongle@seas.upenn.edu}}\, \, 
  Ryan Lucas$^{2}$\, 
  Chen Wang$^{1}$\, 
  Chuhao Chen$^{1}$\, 
  \\[2pt]
  \textbf{Dinesh Jayaraman$^{1}$}\, 
  \textbf{Eric Eaton$^{1}$}\, 
  \textbf{Lingjie Liu$^{1}$}
  \\[2pt]
  $^{1}$University of Pennsylvania \qquad $^{2}$Massachusetts Institute of Technology
}

\begin{document}

\maketitle


~\\

\begin{figure}[hbtp]
        \vspace{-0.5cm}
        \centering
        \includegraphics[width=1.\textwidth, trim={0 0 0 2cm}]{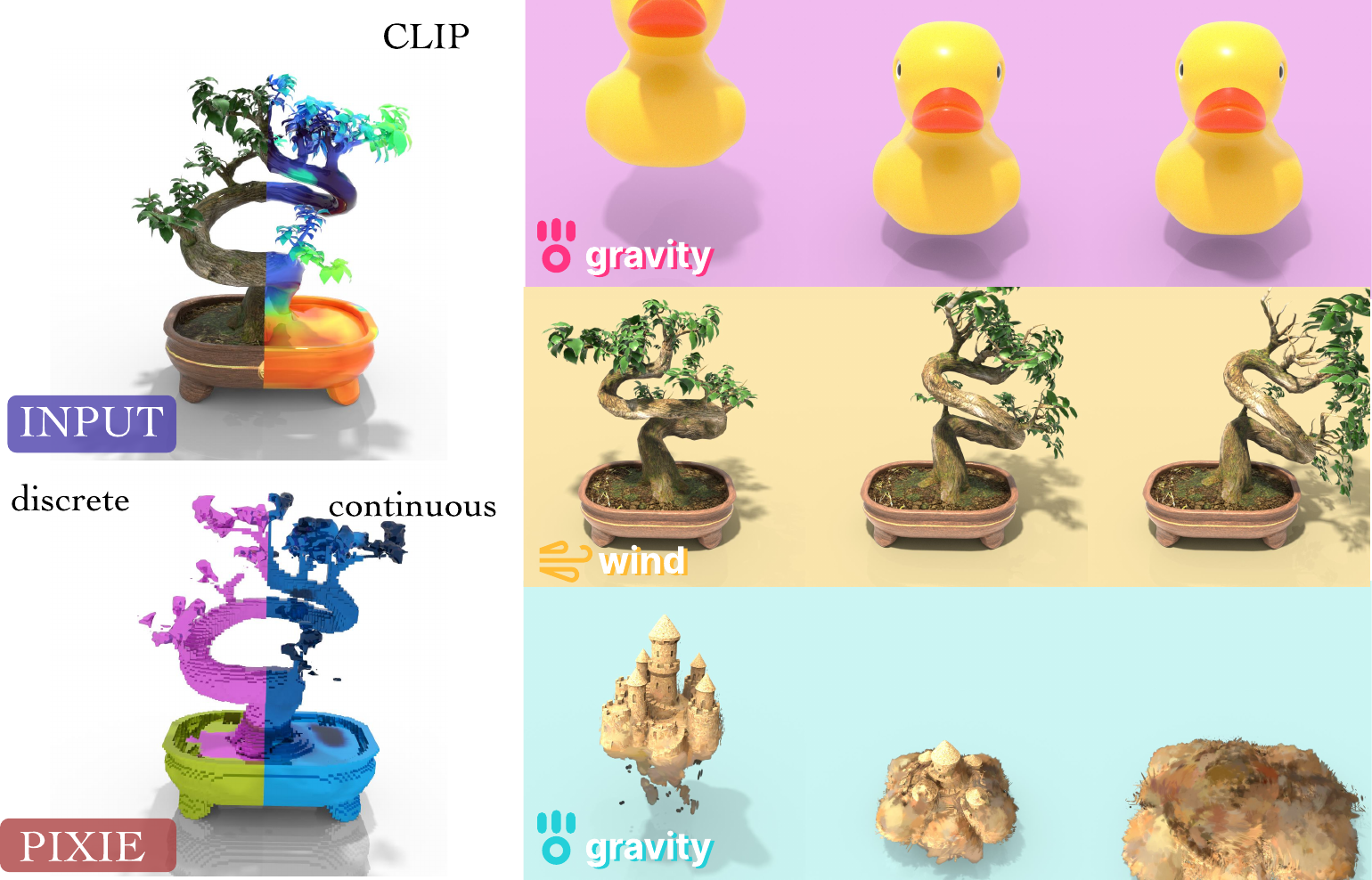}
    \caption{We introduce \method, a novel method for learning simulatable physics of 3D scenes from visual features. Trained on a curated dataset of paired 3D objects and physical material annotations, \method can predict both the discrete material types (e.g., rubber) and continuous values including Young's modulus, Poisson's ratio, and density for a variety of materials, including elastic, plastic, and granular. The predicted material parameters can then be coupled with a learned static 3D model such as Gaussian splats and a physics solver such as the Material Point Method (MPM) to produce realistic 3D simulation under physical forces such as gravity and wind. }
    \label{fig:teaser}
\end{figure}
\input{abstract}

\input{intro}

\input{related_work}


\input{method}

\input{results}

\section{Conclusion and Limitations}
\input{conclusion}

\clearpage
\bibliography{references}

\newpage
\appendix
\input{appendix}

\FloatBarrier  
\clearpage     

\end{document}

%% file: abstract.tex
\begin{abstract}
Inferring the physical properties of 3D scenes from visual information is a critical yet challenging task for creating interactive and realistic virtual worlds. While humans intuitively grasp material characteristics such as elasticity or stiffness, existing methods often rely on slow, per-scene optimization, limiting their generalizability and application. To address this problem, we introduce \method, a novel method that trains a generalizable  neural network to predict physical properties across multiple scenes from 3D visual features purely using supervised losses. Once trained, our feed-forward network can perform fast inference of plausible material fields, which coupled with a learned static scene representation like Gaussian Splatting enables realistic physics simulation under external forces. To facilitate this research, we also collected \dataset, one of the largest known datasets of paired 3D assets and physic material annotations. Extensive evaluations demonstrate that \method is about \gain better and orders of magnitude faster than test-time optimization methods. By leveraging pretrained visual features like CLIP, our method can also zero-shot generalize to real-world scenes despite only ever been trained on synthetic data. \weburl 


\end{abstract}

%% file: intro.tex
\section{Introduction}
\label{sec:intro}

Advances in learning-based scene reconstruction with Neural Radiance Fields \citep{mildenhall2021nerf} and Gaussian Splatting \citep{kerbl20233d} have made it possible to recreate photorealistic 3D geometry and appearance from sparse camera views, with broad applications from immersive content creation to robotics and simulation. However, these approaches focus exclusively on visual appearance---capturing the geometry and colors of a scene while remaining blind to its underlying physical properties.

Yet the world is not merely a static collection of shapes and textures. Objects bend, fold, bounce, and deform according to their material composition and the forces acting upon them. Consequently, there has been a growing body of work that aims to integrate physics into 3D scene modeling \citep{pumarola2020d, ma2023learning, li2023pacnerf, fischer2024sama, feng2023pienerf, xie2023physgaussian, qiu2024feature, guo2024physcomp, lin2025omniphysgs, zhai2024physical, chen2025vid2sim}. Current approaches for acquiring the material properties of the scene generally fall into two categories, each with significant limitations. Some works such as \citep{xie2023physgaussian, guo2024physcomp} require users to manually specify material parameters for the entire scene based on domain knowledge. This manual approach is limited in its application as it places a heavy burden on the user and lacks fine-grained detail. Another line of work aims to automate the material discovery process via test-time optimization. Works including \citep{gradsim, li2023pacnerf, zhong2024springgaus, huang2024dreamphysics, lin2025omniphysgs, zhang2024physdreamer} leverage differentiable physics solvers, iteratively optimizing material fields by comparing simulated outcomes against ground-truth observations or realism scores from video generative models. However, predicting physical parameters for hundreds of thousands of particles from sparse signals (i.e., a single rendering or distillation scalar loss) is an extremely slow and difficult optimization process, often taking hours on a single scene. Furthermore, this heavy per-scene memorization does not generalize: for each new scene, the incredibly slow optimization has to be run from scratch again.

In this paper, we propose a new framework, \method, which unifies geometry, appearance, and physics learning via direct supervised learning. Our approach is inspired by how humans intuitively understand physics: when we see a tree swaying in the wind, we do not memorize the stiffness values for each specific coordinate $(x, y, z)$ -- instead, we learn that objects with tree-like visual features behave in certain ways when forces are applied. This physical understanding from visual cues allows us to anticipate the motion of a different tree or even other vegetation like grass, in an entirely new context. Thus, our insight is to leverage rich 3D visual features such as those distilled from CLIP \citep{radford2021learning} to predict physical materials in a direct supervised and feed-forward way.  Once trained, our model can associate visual patterns (e.g., "if it looks like vegetation") with physical behaviors (e.g., "it should have material properties similar to a tree"), enabling fast inference and generalization across scenes. To facilitate this research, we have curated and labeled \dataset, a dataset of \numobjects paired 3D objects and annotated materials spanning \numclasses semantic classes. We developed a sophisticated multi-step and semi-automatic data labeling process, distilling pretrained models including Gemini \citep{team2023gemini}, CLIP \citep{radford2021learning}, and human prior into the dataset. To our knowledge, this is the largest open-source dataset of paired 3D assets and physical material labels. Trained on \dataset, our feed-forward network can predict material fields that are \gain better and orders of magnitude faster than test-time optimization methods. By leveraging pretrained visual features, \method can also zero-shot generalize to real-world scenes despite only ever being trained on synthetic data.

Our contributions include:
\begin{enumerate}
    \item {\bf Novel Framework for 3D Physics Prediction}: We introduce \method, a unified framework that predicts discrete material types and continuous physical parameters (Young’s modulus, Poisson’s ratio, density) directly from visual features using supervised learning.
\item {\bf \dataset Dataset}: We curate and release \dataset, the largest open-source dataset of 3D objects with physical material annotations (1624 objects, 10 semantic classes).
\item {\bf Fast and Generalizable Inference}: By leveraging pretrained visual features from CLIP and a feed-forward 3D U-Net, \method performs inference orders of magnitude faster than prior test-time optimization approaches, achieving a \gain improvement in realism scores as evaluated by a state-of-the-art vision-language model.
\item {\bf Zero-Shot Generalization to Real Scenes}: Despite being trained solely on synthetic data, \method generalizes to real-world scenes, showing how visual feature distillation can effectively bridge the sim-to-real gap.
\item {\bf Seamless Integration with MPM Solvers}: The predicted material fields can be directly coupled with Gaussian splatting models for realistic physics simulations under applied forces such as wind and gravity, enabling interactive and visually plausible 3D scene animations. 
\end{enumerate}

%% file: related_work.tex
\section{Related Work}
\label{sec:related}


\textbf{2D World Models~~~~} Some early works \citep{bell2015material, bear2021physion} learn to predict material labels on 2D images. Recently, learning forward dynamics from 2D video frames has also been explored extensively. For instance, Google's Genie~\citep{parkerholder2024genie2} trains a next-frame prediction model conditioned on latent actions derived from user inputs, capturing intuitive 2D physics in an unsupervised manner.  While these methods achieve impressive 2D generation and control, they do not explicitly model 3D geometry or a physically grounded world. Other works such as \citep{chen2024unireal, li2024generative} also explore generating or editing images based on learned real-world dynamics. While these methods achieve impressive results in 2D visual synthesis and can imply motion dynamics, they typically do not explicitly model 3D geometry, and only encode physics implicitly via next-frame prediction rather than through explicit material parameters, nor do they infer physically grounded material properties decoupled from appearances. These can lead to problems such as a lack of object permanence or implausible interactions. In contrast, \method directly operates in 3D, predicting explicit physical parameters (e.g., Young's modulus, density) for 3D objects, enabling their integration into 3D physics simulators or neural networks~\cite{wang2025physctrl, mittal2025uniphy} for realistic interaction.

\textbf{Manual Assignment or Assignment of Physics using LLMs~~~~} A number of recent methods have explored combining learned 3D scene representations (e.g., Gaussian splatting) with a physics solver where material parameters are assigned manually or through high-level heuristics. This often involves users specifying material types for the scene~\citep{xie2023physgaussian, abou-chakra2024physically} or using scripted object-to-material dictionaries~\citep{qiu2024feature} or large language and vision-language models~\citep{hsu2024autovfx, chen2025physgen3d, zhai2024physical, le2024articulate, xia2024video2game, li2025wonderplay, cao2025physx} to guide the assignment. 

\textbf{Test-time material optimization using videos~~~~} Other works explore more automatic and principled ways to infer material properties using rendered videos. Some techniques  \citep{gradsim, li2023pacnerf, zhong2024springgaus, jiang2025phystwin, zhang2025particle} optimize material parameters by comparing simulated deformations against ground-truth observations, often requiring ground-truth multi-view videos of objects or ground-truth particle positions under known forces. More recent approaches \citep{huang2024dreamphysics, lin2025omniphysgs, zhang2024physdreamer} use video diffusion models as priors to optimize physics via a motion distillation loss. Notably, these approaches suffer from extremely slow per-scene optimization, often taking hours on a single scene, and do not generalize to new scenes. In stark contrast, \method employs a feed-forward neural network that, once trained, predicts physical parameters in seconds, and can generalize to unseen scenes. A recent work Vid2Sim \citep{chen2025vid2sim} also aims to learn a generalizable material prediction network across scenes. This was done by encoding a front-view video of the object in motion with a foundation video transformer \citep{tong2022videomae} and learning to regress these motion priors into physical parameters. Unlike Vid2Sim, \method does not require videos, relying instead on visual features from static images. Overall, \method can also be used as an informed in conjunction with these test-time methods to further refine predictions.



%% file: method.tex
\section{Method}
\label{sec:method}

\begin{figure}[!t]
        \vspace{-1cm}
        \includegraphics[width=1\textwidth, trim={0 0 0 2cm}]{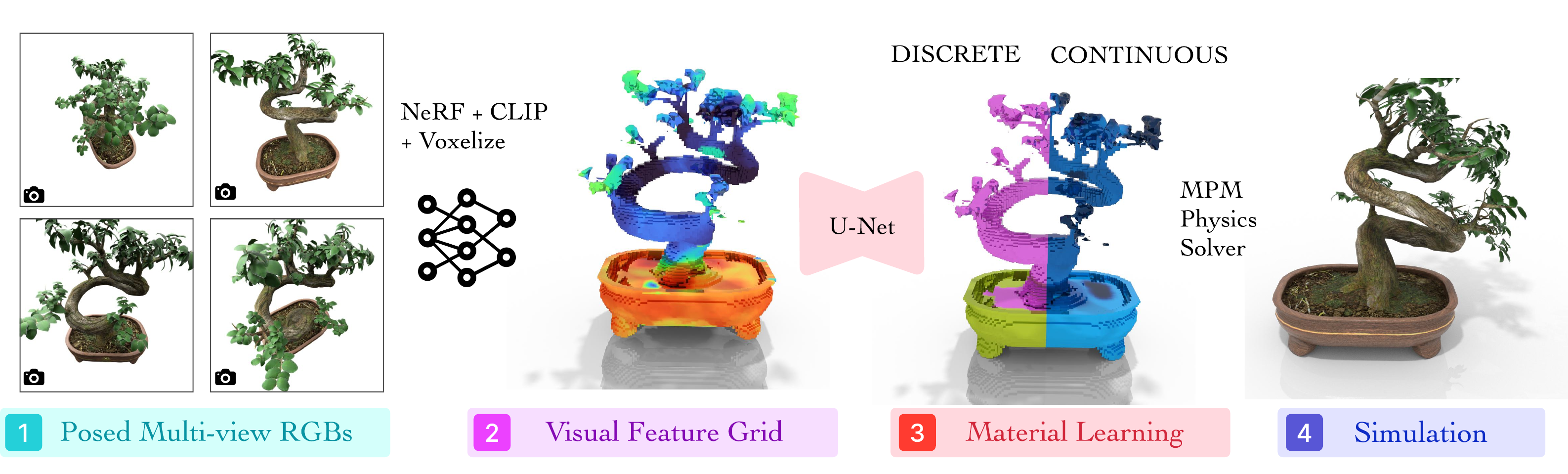}
    \caption{\textbf{Method Overview}. From posed multi-view RGB images of a static scene, \method first reconstructs a 3D model with NeRF and distilled CLIP features \citep{shen2023distilledfeaturefieldsenable}. Then, we voxelize the features into a regular $N \times N \times N \times D$ grid where $N$ is the grid size and $D$ is the CLIP feature dimension. A U-Net neural network \citep{dhariwal2021diffusion} is trained to map the feature grid to the material field $\hat{\mathcal{M}}_G$ which consists of a discrete material model ID and continuous Young's modulus, Poisson's ratio, and density value for each voxel. Coupled with a separately trained Gaussian splatting model, $\hat{\mathcal{M}}_G$ can be used to simulate physics with a physics solver such as MPM.}
    \label{fig:method}
\end{figure}

Our central thesis is that 3D visual appearance provides sufficient information to recover an object's physical parameters. Texture, shading, and shape features captured from multiple calibrated images correlate with physical quantities such as Young's modulus and Poisson's ratio. By learning a mapping from these visual features to material properties, we can augment a volumetric reconstruction model (e.g., Gaussian splatting) with a point-wise material estimate, without requiring force response observations. In Sec.~\ref{sec:method}, we detail our framework, leveraging rich visual priors from CLIP to predict a material field, which can be used by a physics solver to animate objects responding to external forces. To train this model, we curated \dataset, a large dataset of paired 3D assets and material annotations, as detailed in Sec.~\ref{sec:dataset}. Figure~\ref{fig:method} gives an overview of our method.

\vspace{-2mm}
\subsection{\method Physics Learning} \label{sec:method}

{\bf Problem Formulation~~~~}
Formally, the goal is to learn a mapping:
\begin{equation}\label{eq:map}
    f_{\theta} : (\mathcal{I},\Pi)\;\longrightarrow\;\hat{\mathcal{M}}  
\end{equation}
that turns some calibrated RGB images of the static scene \(\mathcal{I}=\{I_k\}_{k=1}^{K}\) and their joint camera specification \(\Pi\) into a
continuous three-dimensional \emph{material field}.  
For every point \(\mathbf{p}\in\mathbb{R}^{3}\) within the scene bounds, the field returns
\[
  \hat{\mathcal{M}}(\mathbf{p}) \;=\;
  \Bigl(
        \hat{\ell}(\mathbf{p}),
        \;
        \hat{E}(\mathbf{p}),
        \;
        \hat{\nu}(\mathbf{p}),
        \;
        \hat{d}(\mathbf{p})
  \Bigr) \enspace,
\]
where
\(\hat{\ell} : \mathbb{R}^{3}\!\to\!\{1,\dots,L\}\) is the discrete material class and
\(\hat{E},\hat{\nu}, d : \mathbb{R}^{3}\!\to\!\mathbb{R}\) are the continuous Young's modulus, Poisson's ratio, and density value respectively. Recall that the discrete material class, also known as the constitutive law, in Material Point Method is a combination of the choices of an expert-defined hyperelastic energy function $\mathcal{E}$ and return mapping $\mathcal{P}$ (Sec.~\ref{sec:prelim}). Learning a point-mapping like this provides a fine-grained material segmentation where for every spatial location we assign both a semantic material label and the physical parameters that characterise that material. Learning the mapping in Eqn.~\eqref{eq:map} directly from 2D images to 3D materials is clearly not simple neither sample efficient. Instead, we leverage a distilled feature field which has rich visual priors to represent the intermediate mapping between 2D images and 3D visual featutes, and then a separate U-Net architecture to compute the mapping between 3D visual features and physical materials. We describe these components below.

\textbf{3D Visual Feature Distillation~~~~}
Recent work on distilled feature fields has shown that dense 2D visual feature embeddings  extracted from foundation models, such as CLIP, based on images can be lifted into 3D, yielding a volumetric representation that is both geometrically accurate and rich in terms of visual and semantic priors \citep{shen2023distilledfeaturefieldsenable}. Here, we also augment the classical NeRF representation \citep{mildenhall2021nerf} to predict a view-independent feature vector in addition to color and density, i.e.,
\[
F_\theta : (\mathbf{x},\mathbf{d}) \mapsto \bigl(\,\mathbf{f}(\mathbf{x}),\; c(\mathbf{x},\mathbf{d}),\; \sigma(\mathbf{x})\bigr) ,
\]
where $c\!\in\!\mathbb{R}^3$ and $\sigma\!\in\!\mathbb{R}_{\ge 0}$ are standard color and radiance NeRF outputs and $\mathbf{f}\!\in\!\mathbb{R}^d$ is a high-dimensional descriptor capturing visual semantics (e.g., object identity or other attributes), which we assume to be view-independent. We supervise color with image RGB and features with per-pixel CLIP embeddings extracted from the training images, using standard volume rendering (App.~\ref{app:ffd}). After training, we voxelize the feature field within known scene bounds to obtain a regular grid  $F_G$ of dimension $N \times N \times N \times D$ grid, where $N=64$ is the grid size and $D=768$ is the CLIP feature dimension, serving as input to our material network.

\textbf{Material Grid Learning~~~~} Our material learning network $f_M$ consists of a feature projector $f_P$ and a U-Net $f_U$. As the CLIP features are very high-dimensional, we learn a feature projector network $f_P$, which consists of three layers of 3D convolution mapping CLIP features $\mathbb{R}^{768}$ to a low-dimensional manifold $\mathbb{R}^{64}$. We then use the U-Net architecture $f_U$ to learn the mapping from the projected feature grid $F_G$ to a material grid $ \hat{\mathcal{M}}_G(\mathbf{p})$, which is a voxelized version of the material field  $\hat{\mathcal{M}}(\mathbf{p})$. The feature projector $f_P$ and U-Net $f_U$ are jointly trained end-to-end via a cross entropy and mean-squared error loss to predict the discrete material classification and the continuous values including Young's modulus, Poisson's ratio and density. More details is in Appendix~\ref{sec:model-arch}.

We found that our voxel grids are very sparse with around 98\% of the voxels being background. Naively trained, the material network $f_M$ would learn to always predict background. Thus, we also separately compute an occupancy mask grid $\mathbb{M} \in \mathbb{R}^N \times \mathbb{R}^N \times \mathbb{R}^N$, constructed by filtering out all voxels whose NeRF densities fall below a threshold $\alpha = 0.01$. The supervised losses---cross entropy and mean squared errors---are only enforced on the occupied voxels. Concretely, the masked supervised loss consists of a discrete cross entropy and continuous mean-squared error loss:
\vspace{-2mm}
\begin{equation}
\label{eq:supervised_loss_multiline}
\begin{split}
\mathcal{L}_{\text{sup}} = \frac{1}{N_{occ}} \sum_{\mathbf{p} \in \mathcal{G}} \mathbb{M}(\mathbf{p}) \Bigl[
    & \lambda \cdot \text{CE}(\hat{\ell}(\mathbf{p}), \ell^{GT}(\mathbf{p})) + (\hat{E}(\mathbf{p}) - E^{GT}(\mathbf{p}))^2 \\[-0.5em]
    & + (\hat{\nu}(\mathbf{p}) - \nu^{GT}(\mathbf{p}))^2  + (\hat{d}(\mathbf{p}) - d^{GT}(\mathbf{p}))^2
\Bigr] \enspace ,
\end{split}
\end{equation}
where $N_{occ} = \sum_{\mathbf{p} \in \mathcal{G}} \mathbb{M}(\mathbf{p})$ is the total number of occupied voxels in the grid, $\hat{\ell}(\mathbf{p})$ and $\ell^{GT}(\mathbf{p})$ are the predicted material class logits and the ground-truth, $CE$ is the cross entropy loss, $\lambda$ is a loss balancing factor, and $E, \nu ,d$ are the Young's modulus, Poisson's ratio and density values, respectively. 

{\bf Physics Simulation~~~~} We use the Material Point Method (MPM) to simulate physics. The MPM solver (Sec.~\ref{subsec:diff_physics}) takes a point cloud of initial particle poses along with predicted material properties, and the external force specification, and simulates the particles' transformations and deformations. Although it is possible to sample particles from a NeRF model (e.g., via Poisson disk sampling \citep{feng2023pienerf}), we have found that it is easier to use a Gaussian Splatting model (Sec.~\ref{sec:prelim}) as each Gaussian can naturally be thought of as a MPM particle \citep{xie2023physgaussian}. Thus, we separately learn a Gaussian splatting model from posed multi-view RGB images. We then transfer the material properties from our predicted material grid into the Gaussian splatting model via nearest neighbor interpolation. 

\begin{figure}[!t]
    \vspace{-1.5cm}
        \centering
        \includegraphics[width=1.\textwidth, trim={0 0cm 0 0cm}, clip]{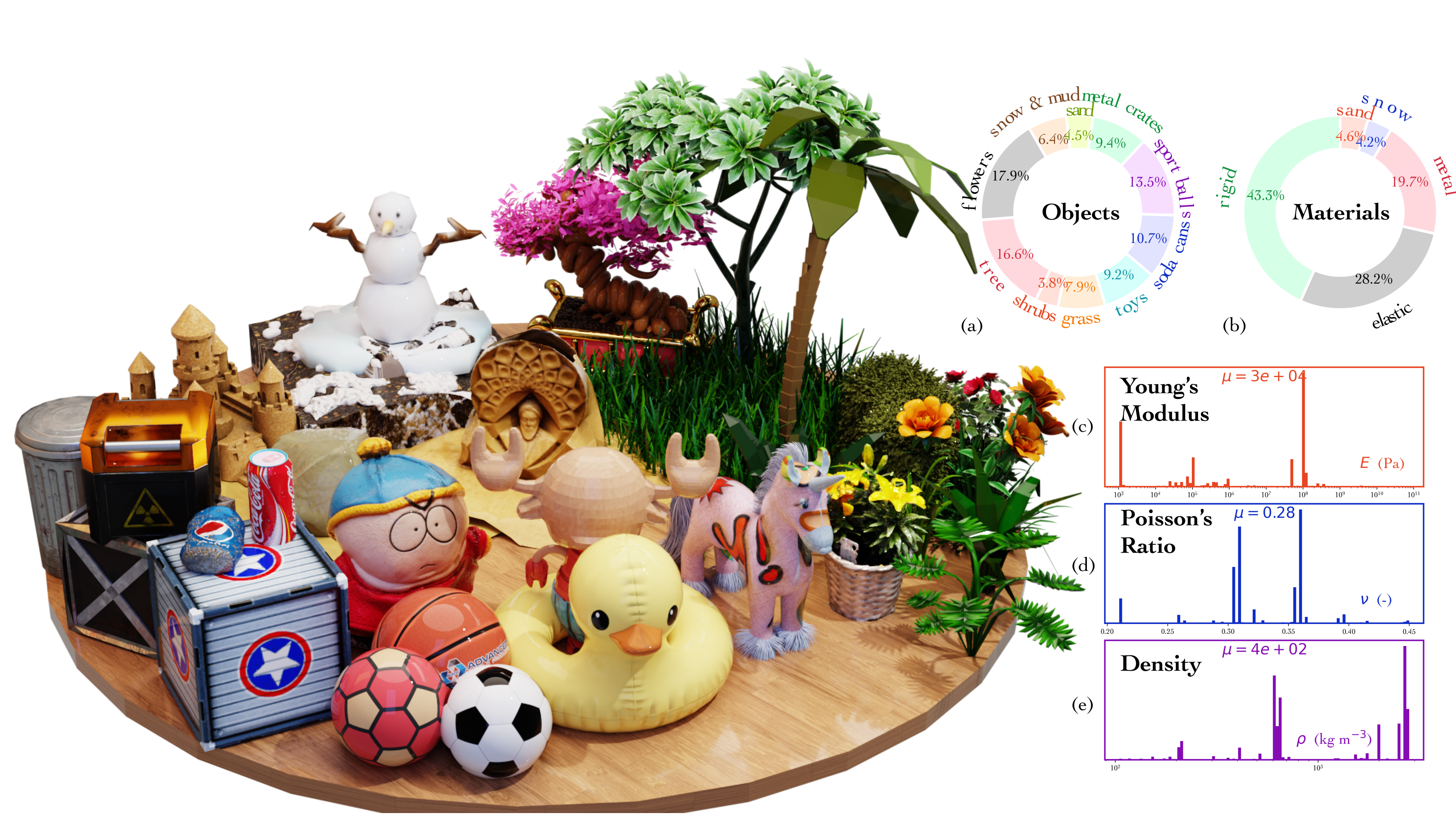}
       \caption{\textbf{\dataset Dataset Overview.}  We collect \numobjects high-quality single-object assets, spanning \numclasses semantic classes (a), and \nummaterials constitutive material types (b). The dataset is annotated with detailed physical properties including spatially varying discrete material types (b), Young's modulus (c), Poisson's ratio (d), and mass density (e). The left figure shows representative examples from the dataset: organic matter (\emph{tree, shrubs, grass, flowers}), deformable toys (\emph{rubber ducks}), sports equipment (\emph{sport balls}), granular media (\emph{sand, snow \& mud}), and hollow containers (\emph{soda cans, metal crates}). }

        \label{fig:dataset}
\end{figure}

\vspace{-3mm}
\subsection{\dataset Dataset} \label{sec:dataset}


We collect one of the largest and highest quality known datasets of diverse objects with annotated physical materials. Our dataset (Fig.~\ref{fig:dataset}) covers \numclasses semantic classes, ranging from organic matter (trees, shrubs, grass, flowers) and granular media (sand, snow and mud) to hollow containers (soda-cans, metal crates), and toys (rubber ducks, sport balls). The dataset is sourced from Objaverse \citep{deitke2022objaverseuniverseannotated3d}, the largest open-source dataset of 3D assets. Since Objaverse objects do not have physical parameter annotations, we develop an semi-automatic multi-stage labeling pipeline leveraging foundation vision-language models i.e., Gemini-2.5-Pro \citep{team2023gemini}, distilled CLIP feature field \citep{kobayashi2022distilledfeaturefields} and manually tuned in-context physics examples. The full details is given in Appendix \ref{sec:dataset_appendix} and \ref{sec:appendix_nerf2physic_ours_data}.

%% file: results.tex
\vspace{-3mm}
\section{Experiments}

 \textbf{Dataset~~~~} We train \method on the \dataset dataset and evaluate on 38 synthetic scenes from the test set and six real-world scene from the NeRF \citep{mildenhall2021nerf}, LERF \citep{kerr2023lerf} datasets, and Spring-Gaus \citep{zhong2024springgaus}.

\textbf{Simulation Details~~~~} We use the material point method (MPM) implementation from PhysGaussian \citep{xie2023physgaussian} as the physics solver. The solver takes a gaussian splatting model augmented with physics where each Gaussian particle also has a discrete material model ID, and continuous Young's modulus, Poisson's ratio, and density values. Each simulation is run for around 50 to 125 frames on a single NVIDIA RTX A6000 GPU. External forces such as gravity and wind are applied to the static scenes as boundary conditions to create physics animations. 

\textbf{Baselines~~~~} We evaluate \method against two recent test-time optimization methods: DreamPhysics \citep{huang2024dreamphysics} and OmniPhysGS \citep{lin2025omniphysgs}, and a LLM method -- NeRF2Physics \citep{zhai2024physical}. DreamPhysics optimizes a Young's modulus field, requiring users to specify other values including material ID, Poisson's ratio, and density. OmniPhysGS, on the other hand, selects a hyperelastic energy density function and a return mapping model, which, in combination, specifies a material ID for each point in the field, requiring other physics parameters to be manually specified. Both methods rely on a user prompt such as "a tree swing in the wind" and a generative video diffusion model to optimize a motion distillation loss. \method, in contrast, infers all discrete and continuous parameters jointly (Fig.~\ref{fig:ours_pred}). NeRF2Physics first captions the scene and query a LLM for all plausible material types (e.g., ``metal") along with the associated continuous values. Then, the material semantic names are associated with 3D points in the CLIP feature field, and physical properties are thus assigned via weighted similarities. This method is similar to our dataset labeling in principle with some crucial differences as detailed in Appendix \ref{sec:dataset_appendix} and \ref{sec:appendix_nerf2physic_ours_data}, allowing \dataset to have much more high-quality labels. \method was trained on 12 NVIDIA RTX A6000 GPUs, each with a batch size of $4$, in one day using the Adam optimizer \citep{kingma2014adam} while prior test-time methods do not require training. For training \method and computing metrics, we apply a log transform to $E$ and $\rho$, and normalize all $\log E, \nu, 
\log \rho$ values to $[-1, 1]$ based on max/min statistics from \dataset.

\textbf{Evaluation Metrics~~~~} We utilize a state-of-the-art vision-language model, Gemini-2.5-Pro \citep{team2023gemini} as the judge. The models are prompted to compare the rendered candidate animations generated using physics parameters predicted by different baselines, and score those videos on a scale from 0 to 5, where a higher score is better. The prompt is in Appendix \ref{sec:vlm_judge}. We also measure the reconstruction quality using PSNR and SSIM metric against the reference videos in the \dataset dataset, which are manually verified by humans for quality control. Other metrics our method optimizes including class accuracy and continuous errors over $E, \nu, \rho$ are also computed. 


\input{table}
\begin{figure}[!t]
\vspace{-1.5cm}
        \centering
        \includegraphics[width=1.\textwidth, trim={0 0cm 0 0cm}, clip]{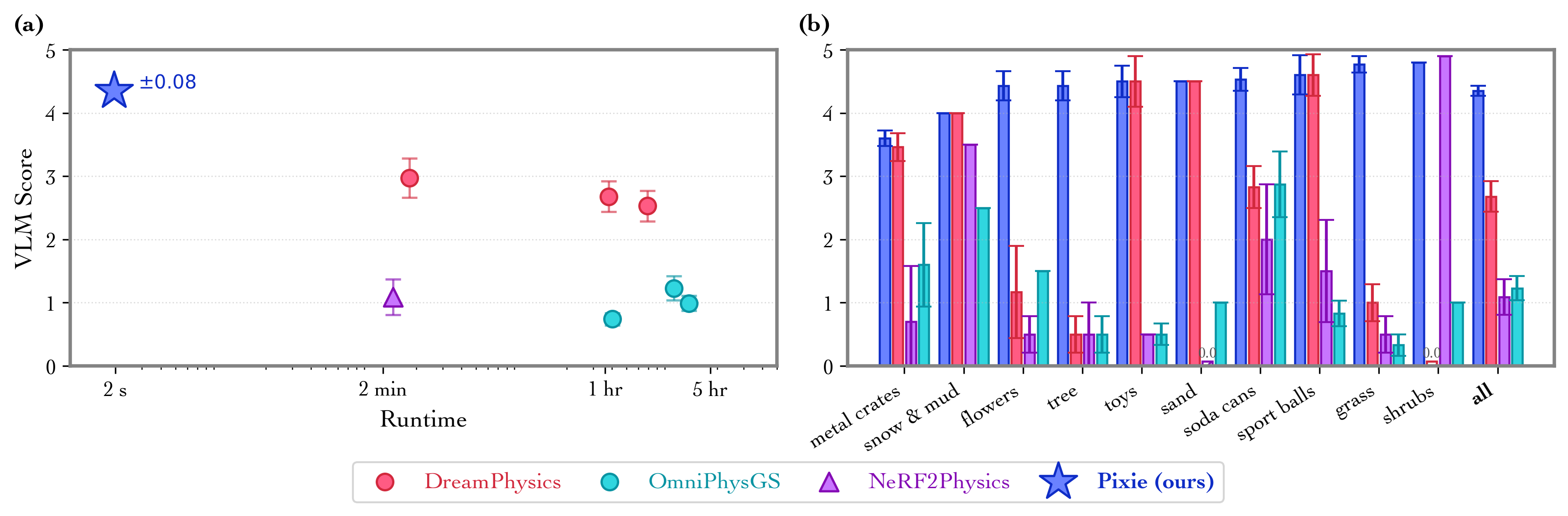}
       \caption{\textbf{Main VLM Results.} (a) \textbf{VLM score versus wall-clock time:} \method is three orders of magnitude faster than previous works while achieving \gain improvement in realism. Test-time optimization methods are run with varying numbers of epochs i.e., $1,25, 50$ for DreamPhysics and $1, 2, 5$ for OmniPhysGS while inference methods are only run once. (b) \textbf{Per-class VLM score:} Our method leads on most object classes. Standard errors are also included.}
        \label{fig:main_comparison}
\end{figure}

\subsection{Synthetic Scene Experiments}

\begin{figure}[!t]
        \vspace{-1.5cm}
        \centering
              \includegraphics[width=1\textwidth, trim={0 0 0 0cm}]{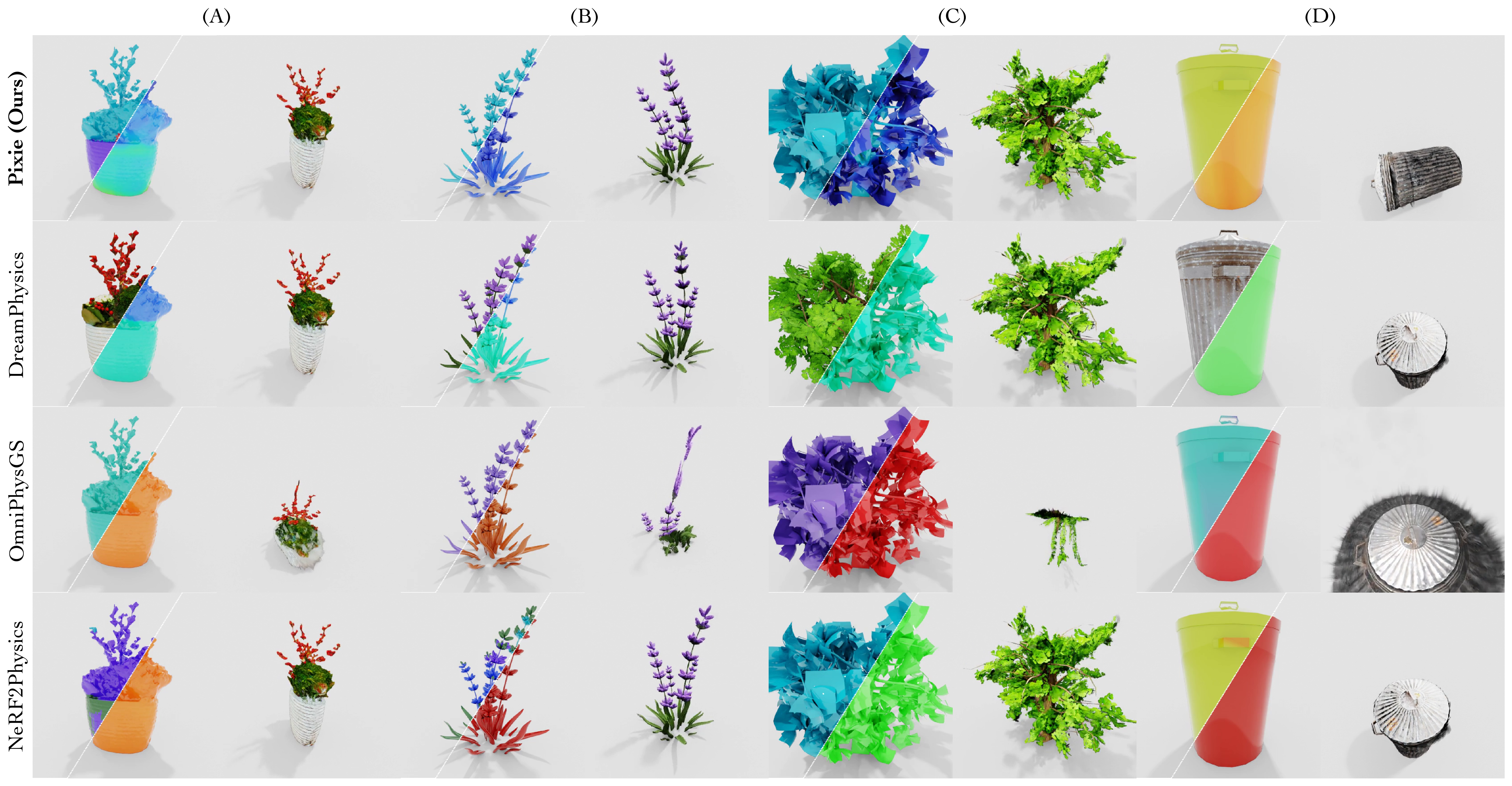}
              \includegraphics[width=1\textwidth, trim={0 0 0 0.5cm}]{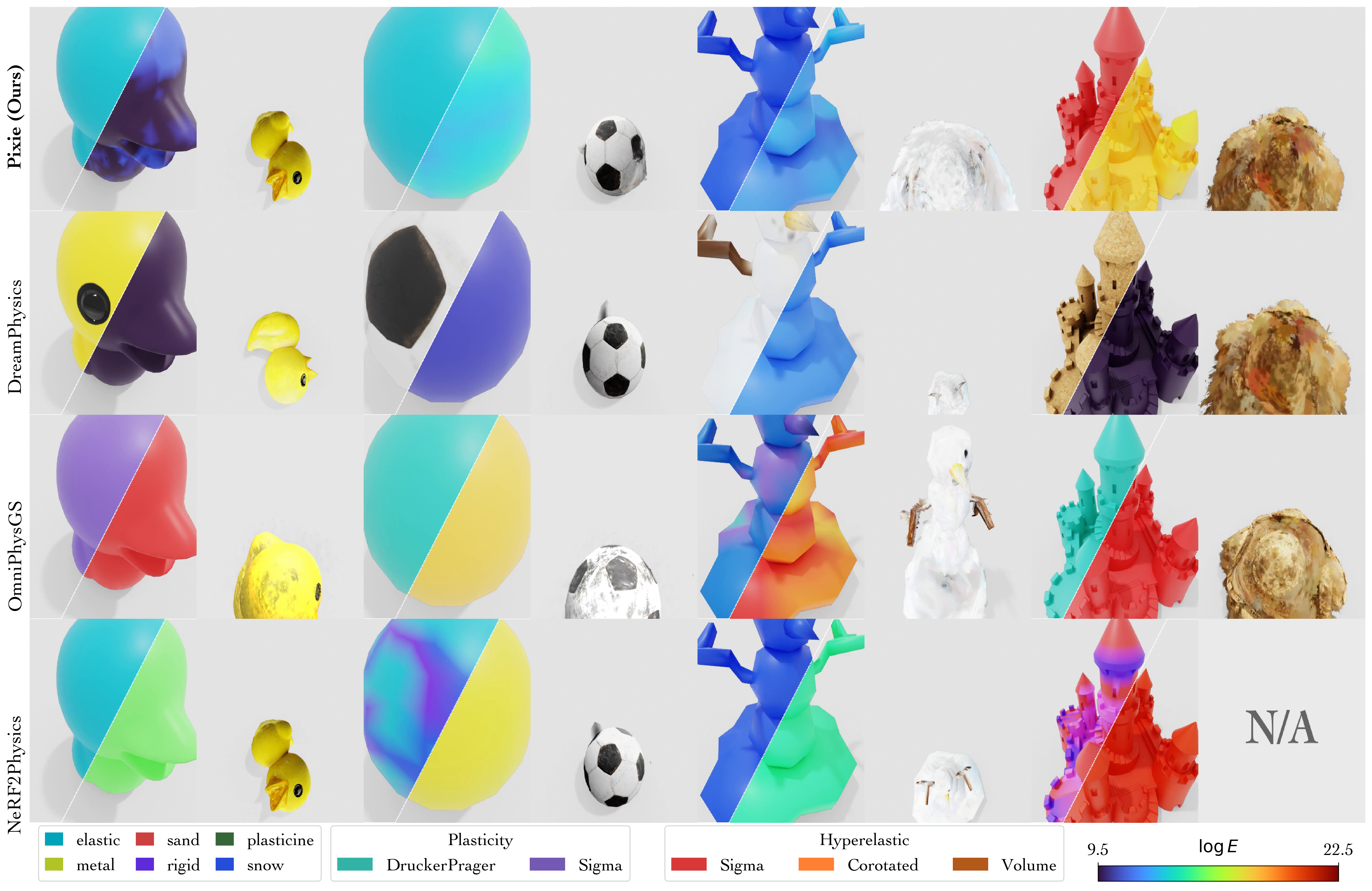}
    \caption{\textbf{Qualitative comparison on synthetic scenes.} 
     We visualized the predicted material class and $E$ predictions (left, right respectively) for \method and Nerf2Physics, $E$ for DreamPhysics (right), and the plasticity and hyperelastic function classes predicted by OmniPhysGS. \method produces stable, physically plausible motion while DreamPhysics remains overly stiff due to inaccurate fine-grained $E$ prediction or too high $E$ (e.g., see tree (C)), OmniPhysGS collapses under load due to unrealistic combination of plasticity and hyperelastic functions, and NeRF2Physics exhibits noisy artifacts. Please see \weburl for the videos.}
    
    
    \label{fig:syn_res}
\end{figure}

Figure~\ref{fig:main_comparison} (a) plots Gemini score versus runtime.  \method achieves a VLM realism score of \textbf{4.35 $\pm$ 0.08} -- a \textbf{1.46-4.39x} improvement over all baselines and tops all other metrics -- while reducing inference time from minutes or hours to \textbf{2 s}. A per-class breakdown in Fig.~\ref{fig:main_comparison} (b) shows our lead in most classes. In Table~\ref{tab:avg-metrics-comparison}, our model improves perceptual metrics such as PSNR and SSIM by $3.6-30.3\%$ and VLM scores by $2.21-4.58$x over prior works.  Figure~\ref{fig:syn_res} visualises eight representative scenes, comparing \method against prior works. DreamPhysics leaves stiff artifacts due to missegmentation or overly high predicted $E$ values, OmniPhysGS collapses under force, and NeRF2Physics introduces high-frequency noise, whereas \method generates smooth, class-consistent motion and segment boundaries. In the appendix, Figure \ref{fig:ours_pred} qualitatively visualizes the physical properties predicted by our network, showing \method's ability to cleanly and accurately recover both discrete and continuous parameters across a diverse sets of objects and continuous value spectrum. In contrast, some prior methods can only recover a subset of parameters like $E$ or material class.


\begin{figure}[!t]
        \vspace{-1.5cm}
        \centering
        \includegraphics[width=1\textwidth, trim={0 0 0 0}]{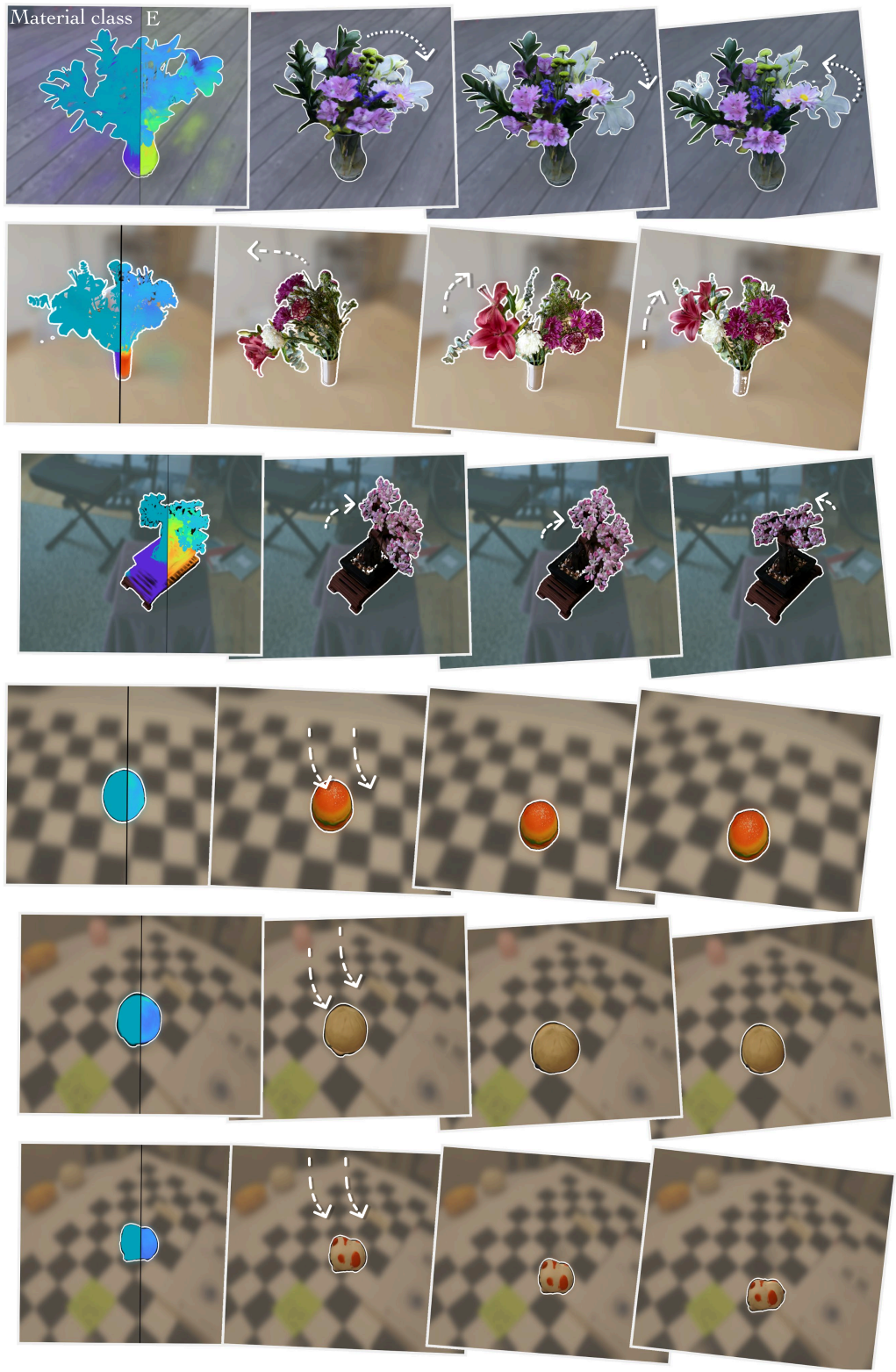}
    \caption{\textbf{\method's Zero-shot Real-scene Generalization.} Trained only on synthetic \dataset, \method can predict plausible physic properties, enabling realistic MPM simulation of real scenes. Here, we visualize the material types (left) and Young's modulus (right) prediction in the first frame, and subsequent frames impacted by a wind force. Please see the videos in our website \weburl.}
    \label{fig:real_res}
\end{figure}



\subsection{Zero-shot Generalization to Real-World Scenes}

Without any real-scene supervision, \method can zero-shot generalize to many real-world scenes as shown in Fig.~\ref{fig:real_res}. For example, our method correctly assigns rigid vase bases and flexible leaves, yielding realistic motion that closely matches human expectation. Our method is surprisingly performant despite significant and non-trivial visual gaps between the training synthetic data versus the out-of-distribution real-world scenes. No other baseline can generalize under this setting.

\subsection{\method's Feature Type Ablation}
Replacing CLIP with RGB or occupancy features drops VLM score by 40-60\,\% and nearly doubles parameter MSE (Table~\ref{tab:avg-metrics-comparison}, rows “Occupancy” and “RGB”). We provide more results in the Appendix. Specifically, we show that the material class prediction also dramatically drops across all classes as shown in Fig.~\ref{fig:ablation_acc}.  Figure~\ref{fig:real_res_ablation} shows the failure modes for real scenes, highlighting RGB and occupancy's struggle to generalize to unseen data as compared to CLIP.

%% file: table.tex
\begin{table}[!t]
\vspace{-1cm}
\centering
\small
\setlength{\tabcolsep}{4pt}
\caption{\textbf{Main Quantitative Results.} We report the average reconstruction quality (PSNR, SSIM) against the reference videos in \dataset, the VLM score, and five other metrics our method optimizes including material accuracy and continuous errors over $E, \nu, \rho$. Standard errors and 95\% CI are also included, and best values are \textbf{bolded}. \method-CLIP is by far the best method across all metrics, achieving 1.62-5.91x improvement in VLM score and 3.6-30.3\% gains in PSNR and SSIM. Our CLIP variant is also notably more accurate than RGB and occupancy features as measured by material class accuracy and average continuous MSE on the test set. While our method simultaneously recovers all physical properties, some prior works only predict a subset, hence $\textit{-}$.}
\vspace{4pt}
\resizebox{\textwidth}{!}{%
\begin{tabular}{lcccccccc}
\toprule
\textbf{Method} & \textbf{PSNR $\uparrow$} & \textbf{SSIM $\uparrow$} & \textbf{VLM $\uparrow$}
& \textbf{Mat.\ Acc.\,$\uparrow$} & \textbf{Avg.\ Cont.\ MSE $\downarrow$}
& \textbf{$\log E$ err $\downarrow$} & \textbf{$\nu$ err $\downarrow$} & \textbf{$\log \rho$ err $\downarrow$}\\
\midrule
\addlinespace[2pt]
\text{DreamPhysics \citep{huang2024dreamphysics}} \\
\quad 1 epoch
    & \meanstd{19.398}{1.090} & \meanstd{0.880}{0.020} & \meanstd{2.97}{0.31}
    & \textit{-} & \textit{-} & \meanstd{2.393}{0.123} & \textit{-} & \textit{-} \\
\quad 25 epochs
    & \meanstd{19.078}{0.939} & \meanstd{0.881}{0.019} & \meanstd{2.68}{0.24}
    & \textit{-} & \textit{-} & \meanstd{1.419}{0.097} & \textit{-} & \textit{-} \\
\quad 50 epochs
    & \meanstd{19.189}{0.980} & \meanstd{0.880}{0.020} & \meanstd{2.53}{0.24}
    & \textit{-} & \textit{-} & \meanstd{1.387}{0.097} & \textit{-} & \textit{-} \\
\midrule
\addlinespace[2pt]
\text{OmniPhysGS \citep{lin2025omniphysgs}} \\
\quad 1 epoch
    & \meanstd{17.907}{0.359} & \meanstd{0.882}{0.007} & \meanstd{0.74}{0.10}
    & \meanstd{0.072}{0.0511} & \textit{-} & \textit{-} & \textit{-} & \textit{-} \\
\quad 2 epochs
    & \meanstd{17.889}{0.372} & \meanstd{0.882}{0.007} & \meanstd{1.23}{0.19}
    & \meanstd{0.109}{0.0704} & \textit{-} & \textit{-} & \textit{-} & \textit{-} \\
\quad 5 epochs
    & \meanstd{17.842}{0.354} & \meanstd{0.883}{0.007} & \meanstd{0.99}{0.12}
    & \meanstd{0.104}{0.0681} & \textit{-} & \textit{-} & \textit{-} & \textit{-} \\
\midrule
\addlinespace[2pt]
NeRF2Physics \citep{zhai2024physical}
    & \meanstd{18.517}{0.644} & \meanstd{0.886}{0.013} & \meanstd{1.09}{0.28}
    & \meanstd{0.274}{0.001} & \meanstd{0.858}{0.109}
    & \meanstd{1.115}{0.165} & \meanstd{0.462}{0.106} & \meanstd{0.997}{0.162} \\
\midrule
\addlinespace[2pt]
\text{\method} \\
\quad Occupancy
    & \meanstd{17.887}{1.524} & \meanstd{0.866}{0.027} & \meanstd{1.76}{0.41}
    & \meanstd{0.643}{0.052} & \meanstd{0.126}{0.012}
    & \meanstd{0.149}{0.023} & \meanstd{0.124}{0.014} & \meanstd{0.105}{0.015} \\
\quad RGB
    & \meanstd{18.652}{2.031} & \meanstd{0.861}{0.035} & \meanstd{2.53}{0.46}
    & \meanstd{0.722}{0.061} & \meanstd{0.106}{0.015}
    & \meanstd{0.196}{0.032} & \meanstd{0.079}{0.012} & \meanstd{0.045}{0.014} \\
\quad \textbf{CLIP (ours)}
    & \textbf{\meanstd{23.256}{2.456}} & \textbf{\meanstd{0.918}{0.023}} & \textbf{\meanstd{4.35}{0.08}}
    & \textbf{\meanstd{0.985}{0.011}} & \textbf{\meanstd{0.056}{0.005}}
    & \textbf{\meanstd{0.022}{0.004}} & \textbf{\meanstd{0.034}{0.006}} & \textbf{\meanstd{0.112}{0.009}} \\
\bottomrule
\end{tabular}}
\label{tab:avg-metrics-comparison}
\vspace{-4pt}
\end{table}

%% file: conclusion.tex
We presented \method, a framework that jointly reconstructs geometry, appearance, and explicit physical material fields from posed RGB images.  By distilling rich CLIP features into 3D and training a feed-forward 3D U-Net with per-voxel material supervision on our new \dataset dataset, \method avoids the expensive test-time optimization required by prior work.  Once trained, it produces full material fields in a few seconds, improving Gemini realism scores by \gain over DreamPhysics and OmniPhysGS while reducing inference time by three orders of magnitude.  \method leverages CLIP's strong visual priors, which enables zero-shot transfer to real scenes, even though it is only trained on synthetic data. The method enables realistic, physically plausible 3D scene animation with off-the-shelf MPM solvers.

\paragraph{Limitations} We take the first step towards learning a supervised 3D model for physical material prediction. Like prior art, our work focuses on single object interaction leaving multi-object scenes for future investigation. Another limitation is that while our UNet predict a point estimate for each voxel, materials in the real-world contain uncertainty that visual information alone cannot resolve (e.g., a tree can be stiff or flexible). A promising extension is to learn a distribution of materials (e.g., using diffusion) instead.


%% file: appendix.tex

\input{preliminaries_appendix}

\input{dataset_appendix}

\input{human_prior}
\input{vlm_judge}
\input{model_arch}

\input{appendix_result}

%% file: preliminaries_appendix.tex
\section{Preliminaries} \label{sec:prelim}
This section briefly reviews foundational concepts in 3D scene representation and physics modeling relevant to our work. 
\subsection{Learned Scene Representation} \label{sub:scene_repr}
Reconstructing 3D scenes from 2D images is commonly achieved by learning a parameterized representation, \(F_\theta\), optimized to render novel views that match observed images \(\{I^{(i)}\}_{i=1}^M\) given camera parameters \(\{\pi^{(i)}\}_{i=1}^M\). This typically involves minimizing a photometric loss:
\[
\min_\theta \sum_{i=1}^{M} \left\| \hat{I}^{(i)}(\theta) - I^{(i)} \right\|_2^2 \enspace,
\]
where \(\hat{I}^{(i)}(\theta)\) is the image rendered from viewpoint $i$. Two prominent representations are Neural Radiance Fields (NeRF) and Gaussian Splatting (GS) models.

{\bf Neural Radiance Fields (NeRF)} \citep{mildenhall2021nerf} model a scene as a continuous function \(F_\theta : (\mathbf{x}, \mathbf{d}) \mapsto (c, \sigma)\), mapping a 3D location \(\mathbf{x}\) and viewing direction \(\mathbf{d}\) to an emitted color \(c\) and volume density \(\sigma\). Images are synthesized using volume rendering, integrating color and density along camera rays. This process' differentiability allows for end-to-end optimization from images.

{\bf Gaussian Splatting (GS)} \citep{kerbl20233d} represents scenes as a collection of 3D Gaussian primitives, each defined by a center \(\mu_i\), covariance \(\Sigma_i\), color \(\mathbf{c}_i\), and opacity \(\alpha_i\). These Gaussians are projected onto the image plane and blended using alpha compositing to render views.

In our work, the principles of neural scene representation, particularly NeRF-like architectures, are leveraged not only for visual reconstruction but also for creating dense 3D visual feature fields. As detailed in Sec.~\ref{sec:method}, we utilize a NeRF-based model to distill 2D image features (e.g., from CLIP) into a volumetric 3D feature grid. This 3D feature representation, \(F_G\), then serves as the primary input to our physics prediction network. For subsequent physics simulation, GS offers a convenient particle-based representation. 

\subsection{3D Visual Feature Distillation Details}
\label{app:ffd}
Following \citep{shen2023distilledfeaturefieldsenable}, we augment the NeRF mapping to produce features $\mathbf{f}$ alongside color $c$ and density $\sigma$:
\[
F_\theta : (\mathbf{x},\mathbf{d}) \mapsto \bigl(\,\mathbf{f}(\mathbf{x}),\; c(\mathbf{x},\mathbf{d}),\; \sigma(\mathbf{x})\bigr) .
\]
Given a camera ray $r(t)=\mathbf{o}+t\mathbf{d}$ passing through pixel $p$, color $C(p)$ and features $F(p)$ are volume-rendered as
\begin{align}
C(p) &= \int_{t_n}^{t_f} T(t)\,\sigma\!\bigl(r(t)\bigr)\, c\!\bigl(r(t),\mathbf{d}\bigr)\,dt, &
F(p) &= \int_{t_n}^{t_f} T(t)\,\sigma\!\bigl(r(t)\bigr)\, f\!\bigl(r(t)\bigr)\,dt ,
\end{align}
where $T(t)=\exp\!\big(-\int_{t_n}^{t}\sigma(r(s))\,ds\big)$ is the accumulated transmittance from the ray origin to depth $t$. At each training iteration, a batch of rays is sampled from the input views. For each ray $r$ (pixel $p$), we enforce that the rendered color $C(p)$ matches the ground-truth pixel RGB $C^*(p)$, while the rendered feature $F(p)$ matches the corresponding CLIP-based feature vector $F^*(p)$ extracted from the image. The loss of the network is:
\begin{align*}
\mathcal{L}
=\sum_{p} \bigl\|C(p)-C^{\ast}(p)\bigr\|_2^{2}
+\lambda_{\text{feat}} 
\sum_{p} \bigl\|F(p)-F^{\ast}(p)\bigr\|_2^{2}
\enspace ;
\end{align*}
the first term enforces color fidelity, while the second aligns the rendered volumetric CLIP features with the dense 2D features extracted from the training images.

From a trained distilled feature field $F_{\theta}$, we obtain a regular feature grid $F_G$ of dimension $N \times N \times N \times D$ grid, where $N=64$ is the grid size and $D=768$ is the CLIP feature dimension. This is done via voxelization using known scene bounds. For our synthetic dataset, we center and normalize all objects within a unit cube.

\subsection{Material Point Method (MPM) for Physics Simulation}
\label{subsec:diff_physics}

To simulate how objects move and deform under applied forces, a physics engine requires knowledge of their material properties. These properties are typically defined within the framework of continuum mechanics, which describes the behavior of materials at a macroscopic level. The fundamental equations of motion (conservation of mass and momentum) are:
\begin{align}
\rho \frac{D \mathbf{v}}{D t} \;&=\; \nabla \cdot \boldsymbol{\sigma} \;+\; \mathbf{f}^{\mathrm{ext}} 
&
\nabla \cdot \mathbf{v} \;&=\; 0 \enspace,
\label{eqn:cont_mech}
\end{align}
where $\rho$ is mass density, $\mathbf{v}$ the velocity field, $\boldsymbol{\sigma}$ the Cauchy stress tensor, and $\mathbf{f}^{\mathrm{ext}}$ any external force (e.g.\ gravity or user interactions). The material-specific \emph{constitutive laws} define how $\boldsymbol{\sigma}$ depends on the local deformation gradient $\mathbf{F}$. For elastic materials, stress depends purely on the recoverable strain; for plastic materials, a yield condition enforces partial “flow” once strain exceeds a threshold.

{\bf Constitutive Laws and Parameters~~~~}
Most continuum simulations separate the constitutive model into two core components:
\begin{equation}
\label{eqn:constitutive_laws}
\begin{aligned}
\mathcal{E}_\mu &: \mathbf{F}^e \;\mapsto\; \mathbf{P},\\
\mathcal{P}_\mu &: \mathbf{F}^{\,e,\mathrm{trial}} \;\mapsto\; \mathbf{F}^{\,e,\mathrm{new}} \enspace,
\end{aligned}
\end{equation}
where $\mathbf{F}^e$ is the \emph{elastic} portion of the deformation gradient, $\mathbf{P}$ is the (First) Piola--Kirchhoff stress, and $\mu$ represents the set of material parameters (e.g.\ Young’s modulus $E$, Poisson's ratio $\nu$, yield stress). The \emph{elastic law} $\mathcal{E}_\mu$ computes stress from the current elastic deformation, while the \emph{return-mapping} $\mathcal{P}_\mu$ projects any “trial” elastic update $\mathbf{F}^{\,e,\mathrm{trial}}$ onto the feasible yield surface if plastic flow is triggered. Typically, the constitutive laws i.e., $\mathcal{E}_\mu$ and $\mathcal{P}_\mu$ are hand-designed by domain experts. The choice of $\mathcal{E}$ and $\mathcal{P}$ jointly define a class of material (e.g., rubber). Within a material class, additional continuous parameters $\mu$ including Young's modulus, Poisson's ratio and density can be specified for a more granular control of the material properties (e.g., stiffness of rubber). In our work, \method jointly predicts the discrete material model and the continuous material parameters.

%% file: dataset_appendix.tex
\section{\dataset Dataset Details} \label{sec:dataset_appendix}
We heavily curate the dataset to a set of \numobjects objects after a multi-stage filter that removes multi-object scenes, missing textures, duplicated assets, and objects whose material labeling is  either ambiguous or physically implausible. The process is semi-automatic with a VLM-driven multi-stage pipeline while still imparting substantial human prior and labor. \rebutal{We manually tune the physics parameter ranges for each semantic class (e.g., ``tree", ``rubber toy") and 3D segmentation query terms, and provide these as in-context examples for the VLM to align them with human's physical understanding.}

 First, we define some object class (e.g., ``tree") and some alternative query terms (e.g., ``ficus, fern, evergreen etc"). We then use a sentence transformer model \citep{wang2020minilm} to compute the cosine similarity between the search terms and the name of each Objaverse object. We select $k=500$ objects with the highest similarity score for each class, creating an initial candidate pool. However, since Objaverse objects vary greatly in asset quality, lighting conditions, and some scenes contain multiple objects which are not suitable for our material learning, an additional filtering step is needed. The Gemini VLM is prompted to filter out low-quality or unsuitable scenes. A distilled NeRF model is fitted to each object. Then, the VLM is provided five multi-view RGB images of an object, and prompted to provide a list of the object's semantic parts along with associated material class and ranges for continuous values. The ranges such as $E \in \{1e4, 1e5\}$ allow us to simulate a wider range of dynamics from flexible to more rigid trees. The VLM is also prompted to specify a list of constraints such as to ensure that the leaf's density is lower than the trunk's. We then sample the continuous values from the VLM's specified ranges subject to the constraint via rejection sampling. The semantic parts (e.g., ``pot") are used with the CLIP distilled feature field to compute a 3D semantic segmentation of the object into parts, and the sampled material properties are applied uniformly to all points within a part. This ground-truth material and feature fields are then voxelized into regular grids for use in supervised learning by the \method framework.



The following sections provide more details on each step of our semi-automatic labeling process.

\input{prompt_objaverse_selection}
\subsection{Object Selection from Objaverse}
We use the \href{https://huggingface.co/sentence-transformers/all-MiniLM-L6-v2}{all-MiniLM-L6-v2} \citep{wang2020minilm} sentence transformer to compute the cosine similarity between an objaverse asset's name and some search terms for each object class. The search terms are in Fig.~\ref{fig:objaverse_selection}. The top $k=500$ objects with the highest similarity score are selected for each class.

\subsection{Object Filtering}
\begin{figure}
        \vspace{-1cm}
        \centering
        \includegraphics[width=1\textwidth, trim={0 0 0 0}]{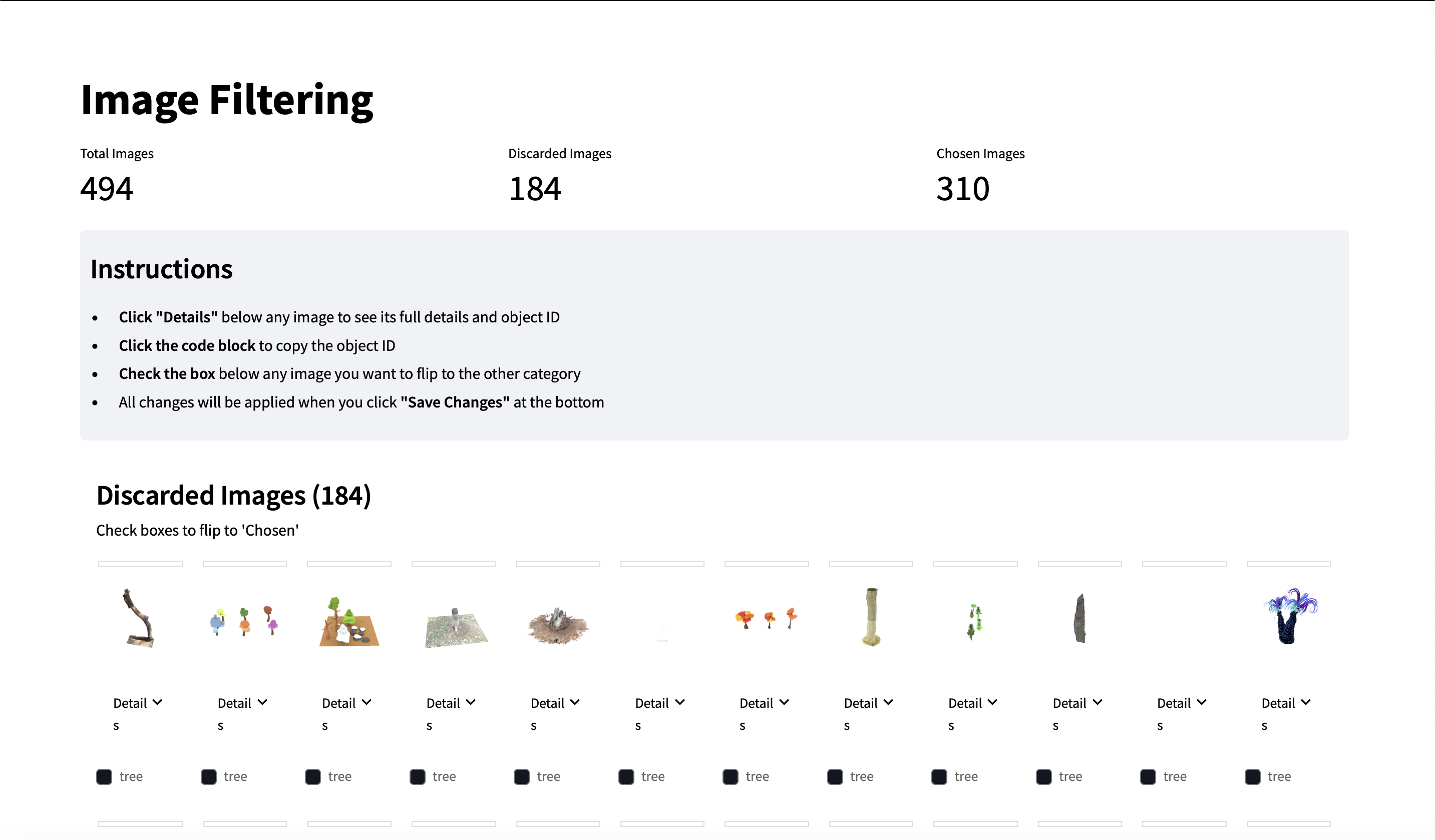}
        \includegraphics[width=1\textwidth, trim={0 0 0 0}]{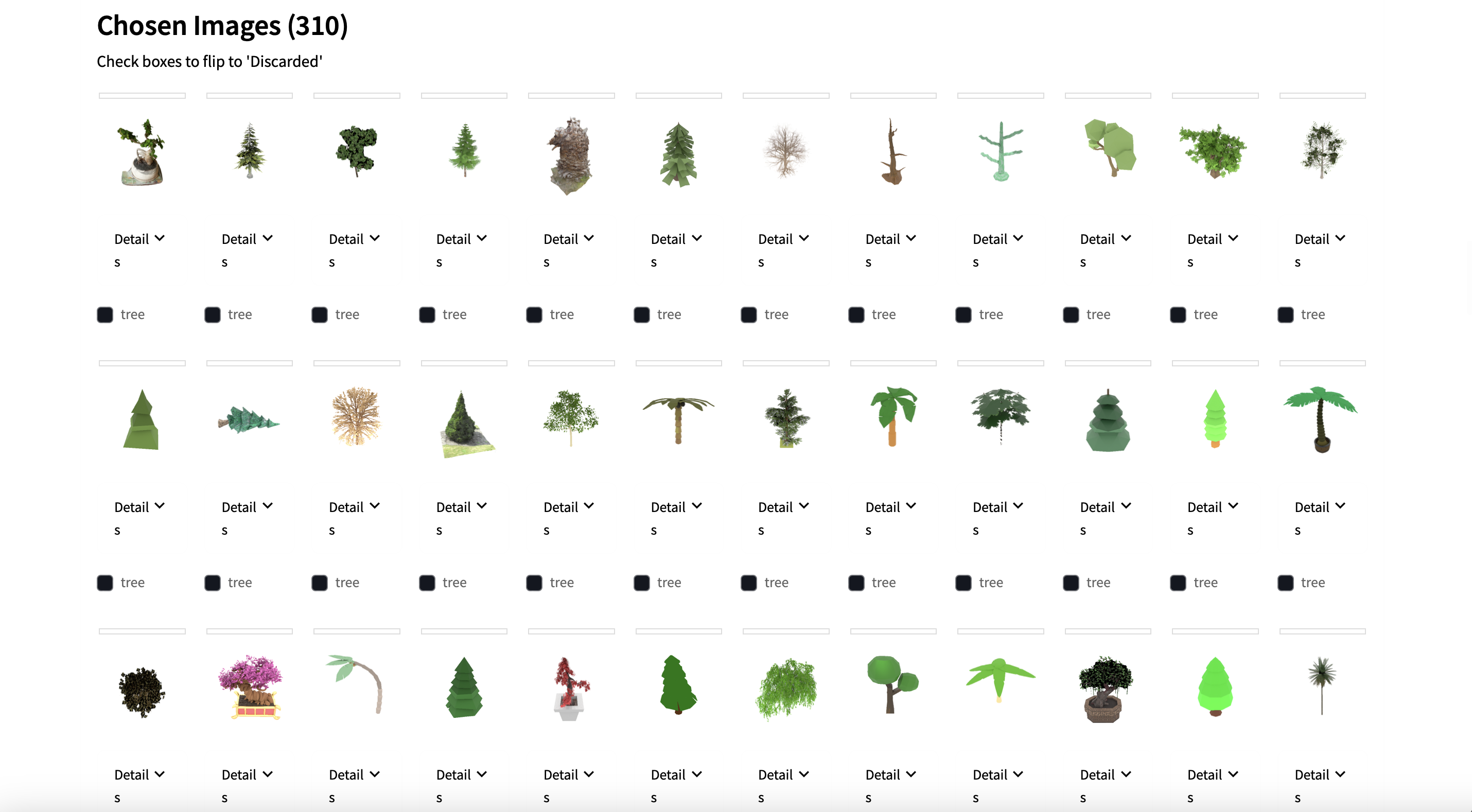}
    \caption{\textbf{Manual correction for object filtering.} The web interface for quickly inspecting and manually correcting VLM results.}
    \label{fig:manual_filtering}
\end{figure}

Next, we prompt Gemini to filter out low-quality assets. The system instruction is given in Fig.~\ref{fig:object_filtering_prompt}. Then, a human quickly scans through the VLM results organized in our web interface as shown in Fig.~\ref{fig:manual_filtering} to correct any mistakes.
\input{prompt_object_filtering}

\subsection{CLIP-Driven 3D Semantic Segmentation}
\begin{figure}
        \vspace{-1cm}
        \centering
        \includegraphics[width=1\textwidth, trim={0 0 0 0}]{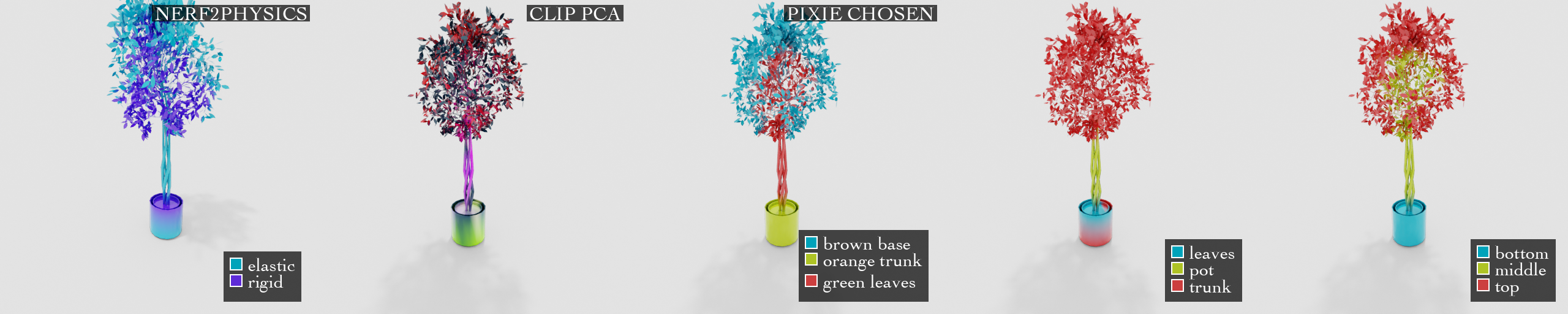}\vspace{-0.15em}\\[-0.15em]\includegraphics[width=1\textwidth, trim={0 0 0 0}]{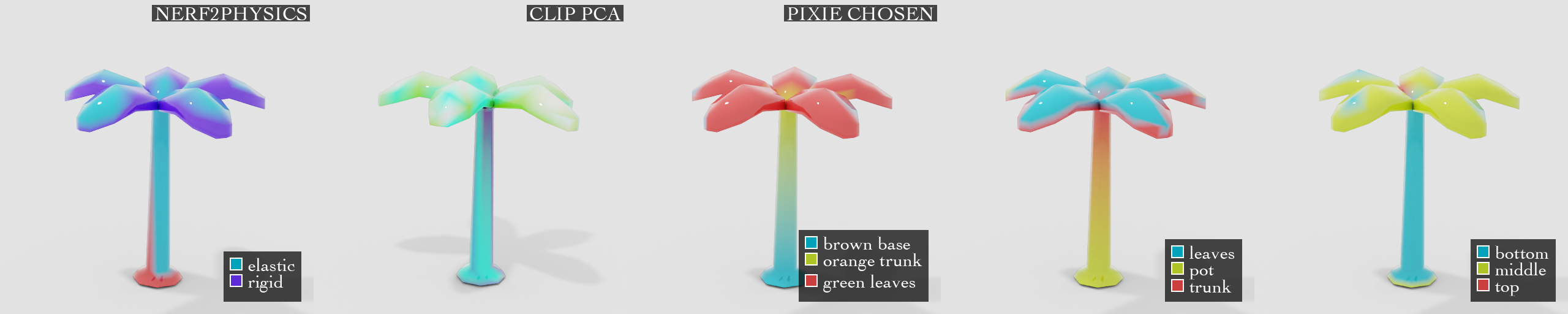}

    \caption{\textbf{CLIP Semantic Segmentation.} CLIP features can be noisy for various objects and different text queries vary greatly in segmentation quality. Thus, we prompt a VLM actor to generate several candidate queries for each object, render all candidates, and prompt another VLM critic to select the best query terms from the rendered 3D segmentation images. Some candidates are provided and proposals chosen by the critic are highlighted. Note that a high-performing query proposal (e.g., ``leaves,pot,trunk") in one object is not necessary high-performant in another. The PCA visualization of the CLIP feature fields is also provided.}
    \label{fig:clip_seg}
\end{figure}

From a distilled CLIP feature field of the object \citep{shen2023distilledfeaturefieldsenable}, we can perform 3D semantic segmentation by providing a list of the object's parts (e.g., ``pot, trunk, leaves"). These query terms are used to compute the cosine similarity between each CLIP feature at a given 3D coordinate against the terms, and the part with highest similarity is assigned to that point. The choices of query terms (e.g., ``pot, trunk, leaves" vs ``base, stem, leaf") greatly affect the segmentation quality, and is not obvious. A high-performing query list in one object is not guaranteed to yield high performance in another object, e.g., see Fig.~\ref{fig:clip_seg}. Thus, we prompt a VLM actor to generate several candidate queries for each object, render all candidates, and prompt another VLM critic to select the best query terms from the rendered 3D segmentation images, as detailed Sec.~\ref{sub:vlm_actor_critic}.

\subsection{VLM Actor-Critic Labeling} \label{sub:vlm_actor_critic}
\begin{figure}
        \centering
        \includegraphics[width=1\textwidth, trim={0 0 0 0}]{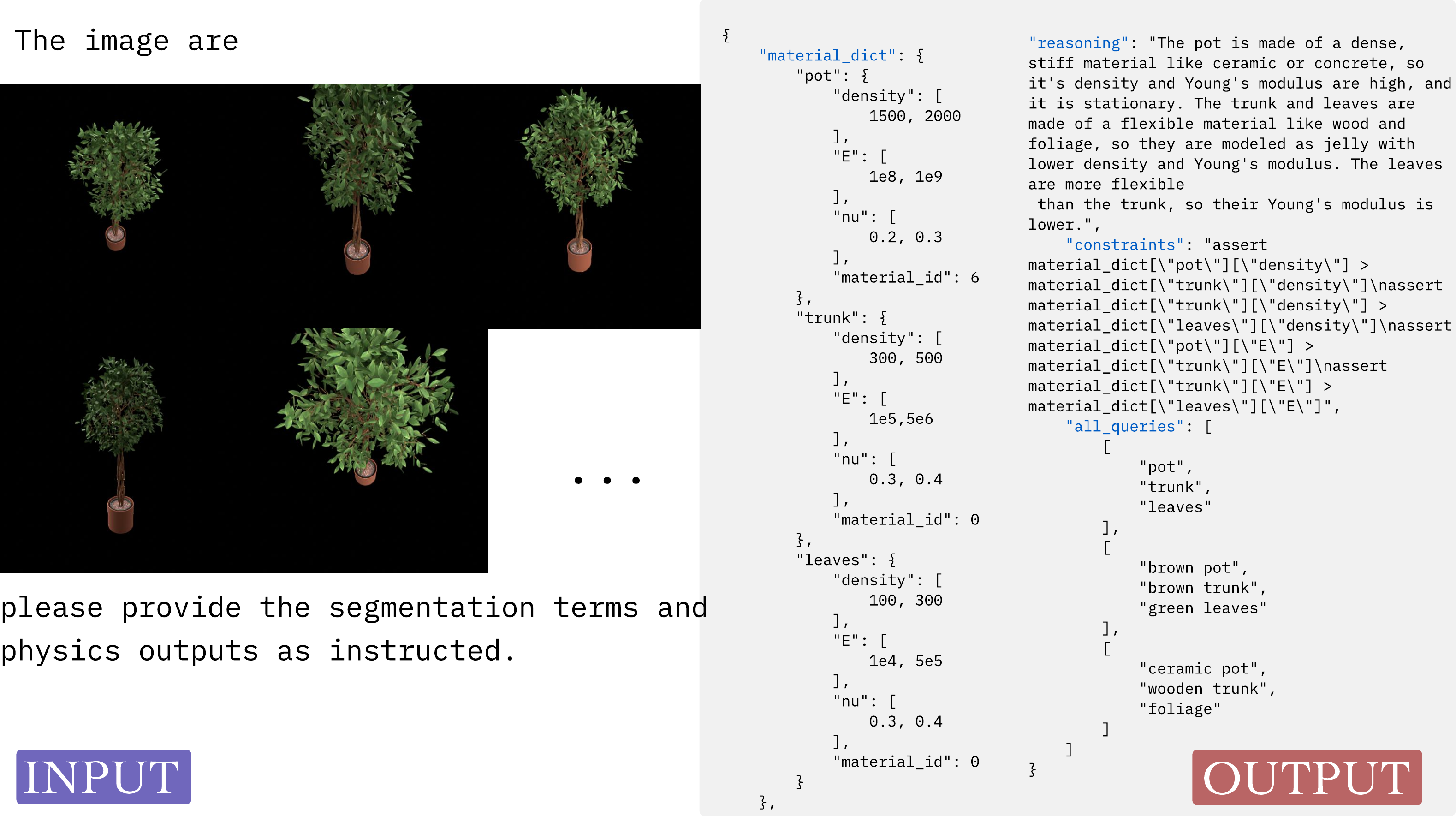}
    \caption{\textbf{VLM Actor's Physics and Segmentation Proposal.}}
    \label{fig:vlm_actor}
\end{figure}

Current VLMs might not have robust physical understanding for generating high-quality labels for \dataset zeroshot. Thus, we first manually tune the physic parameters for each semantic object class (e.g., ``tree", ``rubber toy"). A condensed version of these examples is provided in Fig.~\ref{fig:incontext_physics_toy_examples}. We also provide examples of different search terms (e.g., ``pot, trunk, leaves" vs ``base, stem, leaf"). These in-context examples are provided to a VLM actor that simultaneously proposes physics parameters and semantic segmentatic queries for that object from multi-view images of that object as illustrated in Fig.~\ref{fig:vlm_actor}. The full system prompt for the VLM is provided in  Fig.~\ref{fig:vlm_actor_system_prompt} and the full in-context examples in Listing~\ref{lst:incontext_examples}. We render an image representing 3D semantic segmentation masks for each query proposal as shown in Fig.~\ref{fig:clip_seg}. A VLM critic is then prompted to select the best segmentation queries from the rendered images. The critic's system prompt is provided in Fig.~\ref{fig:vlm_seg_critic_prompt}.

Additionally, materials in the real-world contain uncertainty that visual information alone cannot resolve (e.g., a tree can range from stiff to flexible). Thus, instead of specifying one physics parameter per part, we prompt the VLM actor to output a plausible range (e.g., $E \in \{1e4, 1e5\}$ see Fig.~\ref{fig:vlm_actor}, \ref{fig:incontext_physics_toy_examples}). We then sample a value uniformly from each range to build our training dataset. To further ensure that the sampled values are consistent, the VLM is also prompted to specify a list of constraints (e.g., the density of leaves must be lower than that of the trunk). Rejection sampling is used to ensure that the final dataset respects the constraints. 

\input{prompt_incontext_toy}
\input{prompt_vlm_seg}

\input{prompt_vlm_critic}

\input{prompt_incontext_physics}

%% file: prompt_objaverse_selection.tex
\begin{figure}
    \centering
\begin{tcolorbox}[top=2pt,bottom=2pt,width=\linewidth,boxrule=1pt]{
\begin{lstlisting}[style=promptstyle]
§\textbf{tree}§: tree, ficus, fern, oak tree, pine tree, evergreen, palm tree, maple tree, bonsai tree
§\textbf{flowers}§: flower, bouquet, rose, tulip, daisy, lily, sunflower, orchid, flower arrangement, flowering plant, garden flowers, wildflowers, floral
§\textbf{rubber\_ducks\_and\_toys}§: rubber duck, bath toy, rubber toy, toy duck, squeaky toy, floating toy, plastic duck, children's bath toy, yellow duck toy, rubber animal toy
§\textbf{soda\_cans}§: soda can, aluminum can, beverage can, cola can, soft drink can, metal can, canned drink, pop can, fizzy drink can
§\textbf{sport\_balls}§: basketball, soccer ball, football, tennis ball, baseball, volleyball, golf ball, rugby ball, ping pong ball, cricket ball, bowling ball, beach ball, sports ball
§\textbf{sand}§: sand, beach sand, desert sand, sandy terrain, sand pile, sand dune, sandpit, sand box, sand texture, grainy sand
§\textbf{shrubs}§: shrub, bush, hedge, ornamental bush, garden shrub, boxwood, flowering bush, evergreen shrub, decorative plant, landscaping shrub
§\textbf{metal\_crates}§: metal crate, steel box, metal container, shipping crate, metal storage box, industrial container, metal chest, storage crate, metallic box
§\textbf{grass}§: grass, lawn, turf, grassland, meadow, grassy field, green grass, grass patch, tall grass, wild grass, pasture
§\textbf{snow\_and\_mud}§: snow, mud, snowy ground, muddy ground, wet mud, fresh snow, packed snow, snowy terrain, muddy terrain, snow patch, mud puddle, snowdrift, muddy path, snowy surface, muddy surface, slush, wet snow, dirty snow, muddy water, snowy landscape
\end{lstlisting}
}
\end{tcolorbox}
\caption{\textbf{Objaverse Class Selection Keywords.} The keywords for matching a semantic class with an objaverse asset's name.}
\label{fig:objaverse_selection}
\end{figure}

%% file: prompt_object_filtering.tex
\begin{figure}
    \centering
\begin{tcolorbox}[top=2pt,bottom=2pt,width=\linewidth,boxrule=1pt]{
\begin{lstlisting}[style=promptstyle]
We need to select some images of the classes: {§\textcolor{blue}{class\_name}§}. This class includes objects like {§\textcolor{blue}{search\_terms}§}. We will provide you some images rendered from the 3D model. You need to either return True or False. Return False to reject the image as inappropriate for the video game development. Some common reasons for rejection:
  - The image doesn't clearly depict the object class
  - The image is too dark or too bright or too blurry or has some other low quality.
    Remember, we want high-quality training data.
  - The image contains other things in addition to the object. 
REMEMBER, we only want images that depict cleanly ONE SINGLE OBJECT belonging to one of the classes. But you also need to use your common sense and best judgement. For example, for a class like "flowers", the object might include a vase of flowers (you rarely see a single flower in the wild). So you should return True in this case.
  - We do want diversity in our dataset collection. So even if the texture of the object is a bit unusual, as long as you can recognize it as belonging to the class / search terms, you should return True. Only remove low-quality assets.

The return format is:
```json
{
  "is_appropriate": true (or false),
  "reason": "reason for the decision"
}
```
We'll be using the 3d models to learn physic parameters like material and young modulus to simulate the physics of the object. E.g., the tree swaying in the wind or thing being dropped from a height. Therefore, you need to decide if the image depicts an object that is likely to be used in a physics simulation.
\end{lstlisting}
}
\end{tcolorbox}
\caption{\textbf{Object Filtering System Prompt.} Prompt for VLM to filter out low-quality assets.}
\label{fig:object_filtering_prompt}
\end{figure}

%% file: prompt_incontext_toy.tex
\begin{figure}
    \centering
\begin{tcolorbox}[top=2pt,bottom=2pt,width=\linewidth,boxrule=1pt]{
\begin{lstlisting}[style=promptstyle]
§\textcolor{blue}{tree}§: 
  pot: {density: 400, E: 2e8, nu: 0.4, material: "rigid"}
  trunk: {density: 400, E: 2e6, nu: 0.4, material: "elastic"}
  leaves: {density: 200, E: 2e4, nu: 0.4, material: "elastic"}

§\textcolor{blue}{flowers}§:
  vase: {density: 500, E: 1e6, nu: 0.3, material: "rigid"}
  flowers: {density: 100, E: 1e4, nu: 0.4, material: "elastic"}

§\textcolor{blue}{shrub}§:
  stems: {density: 300, E: 1e5, nu: 0.35, material: "elastic"}
  twigs: {density: 250, E: 6e4, nu: 0.38, material: "elastic"}
  foliage: {density: 150, E: 2e4, nu: 0.40, material: "elastic"}

§\textcolor{blue}{grass}§:
  blades: {density: 80, E: 1e4, nu: 0.45, material: "elastic"}
  soil (if visible): {density: 1200, E: 5e5, nu: 0.30, material: "rigid"}

§\textcolor{blue}{rubber\_ducks\_and\_toys}§:
  toy: {density: [80, 150], E: [3e4, 5e4], nu: [0.4, 0.45], material: "elastic"}

§\textcolor{blue}{sport\_balls}§:
  ball: {density: [80, 150], E: [3e4, 5e4], nu: [0.4, 0.45], material: "elastic"}

§\textcolor{blue}{soda\_cans}§:
  can: {density: [2600, 2800], E: [5e10, 8e10], nu: [0.25, 0.35], material: "metal"}

§\textcolor{blue}{metal\_crates}§:
  crate: {density: [2500, 2900], E: [8e7, 1.2e8], nu: [0.25, 0.35], material: "metal"}

§\textcolor{blue}{sand}§:
  sand: {density: [1800, 2200], E: [4e7, 6e7], nu: [0.25, 0.35], material: "sand"}

§\textcolor{blue}{jello\_block}§:
  jello: {density: [40, 60], E: [800, 1200], nu: [0.25, 0.35], material: "elastic"}

§\textcolor{blue}{snow\_and\_mud}§:
  snow_and_mud: {density: [2000, 3000], E: [8e4, 1.2e5], nu: [0.15, 0.25], material: "snow"}
\end{lstlisting}
}
\end{tcolorbox}
\caption{\textbf{In-Context Physics Condensed Examples.} Material properties for each object class used in the VLM prompting. Density is in kg/m³, E (Young's Modulus) is in Pa, nu (Poisson's ratio) is dimensionless.}
\label{fig:incontext_physics_toy_examples}
\end{figure} 

%% file: prompt_vlm_seg.tex
\begin{figure}[t]
\vspace{-1.4cm}
    \centering
\begin{tcolorbox}[top=1pt,bottom=1pt,width=\linewidth,boxrule=1pt]{
\begin{lstlisting}[style=promptstyle]
We are trying to label a 3D object with physical properties. The physical properties are:
    - Density
    - Young's Modulus
    - Poisson's Ratio
    - Material model

where the material model is one of the following: \{§\textcolor{blue}{material\_list\_str}§\}
We have an automatic semantic segmentation model that can segment the object into different parts. We'll assume that each part has the same material model.

Your job is to come up with the part query to pass to the semantic segmentation model, and the associated material properties for each part.
    \{§\textcolor{blue}{special\_notes}§\}
For example, for a \{§\textcolor{blue}{class\_name\_for\_example}§\}, the return is

    ```json
    \{§\textcolor{blue}{example\_material\_dict\_str}§\}
    ```
    \{§\textcolor{blue}{example\_explanation}§\}
Note that there are many different valid values for the material properties including E, nu, and density that would influence how the object behaves. Thus, instead of actual values, you should return a range of values like "E": [2e4, 2e6]. Also, provide reasoning and constraints on the values when appropriate.

So the output should be a json with the following format:

    ```json
    \{\{
        "material\_dict": \{\{ ... similar to example\_dict with ranges ... \}\},
        "reasoning": "...",
        "constraints": "...",
        "all\_queries": "..."
    \}\}
    ```
Remember to write constraints in the form of python code. For example,
```python
    \{§\textcolor{blue}{example\_constraints\_str}§\}
```
Note that you've been asked to generate a material range so `material\_dict["leaves"]["density"]` is a range of values. But for the purpose of the constraints writing, you can assume that the material\_dict["leaves"]["density"] is a single value, and generate the python code similar to the example above. This is important because we will first sample a value from the range, then invoke your constraints code. So instead of writing something like
```python
    assert material\_dict["leaves"]["density"][0] ...
```
you must write something like
```python
    assert material\_dict["leaves"]["density"] ...
```
Note that the correct code doesn't have the bracket because `material\_dict["leaves"]["density"]` will be already reduced to a single value by our sampler.
You will be provided with images of the object from different views or a single view. Please try your best to come up with appropriate part queries as well. For example, if the object doesn't have visible trunk or pot, then you should NOT include them in the material\_dict. Only segment parts that are visible in the image.
Also, because our CLIP segmentation model is not perfect, you should come up with alternative queries as well including the original queries in the all\_queries list. For example,
    ```json
    \{§\textcolor{blue}{example\_all\_queries\_str}§\}
    ```
In total, you need to provide \{§\textcolor{blue}{num\_alternative\_queries}§\} alternative queries. 
Tips:
\{§\textcolor{blue}{tips\_str}§\}
- Make sure that each element in the `all\_queries` list is in the exact same order as the material\_dict keys.
\end{lstlisting}
}
\end{tcolorbox}

\caption{\textbf{VLM Actor System Prompt.}}
\label{fig:vlm_actor_system_prompt}
\end{figure}

%% file: prompt_vlm_critic.tex
\begin{figure}
    \centering
\begin{tcolorbox}[top=2pt,bottom=2pt,width=\linewidth,boxrule=1pt]{
\begin{lstlisting}[style=promptstyle]
You are a segmentation quality critic. Your task is to evaluate the quality of segmentation results produced by a CLIP-based segmentation model.

You will be shown:
1. A set of original RGB images of a 3D object from different views
2. Segmentation results for different part queries

Your job is to:
1. Evaluate each segmentation query based on how well it separates the object into meaningful parts
2. Score each query on a scale of 1-10 (10 being perfect)
3. Provide reasoning for your scores
4. Suggest improvements to the queries if needed

Consider the following factors in your evaluation:
- Does the segmentation properly separate the object into distinct, semantically meaningful parts?
- Are the boundaries of the segments accurate and clean?
- Is any important part of the object missed or incorrectly segmented?
- IMPORTANT: note that our imperfect CLIP segmentation model is heavily dependent on the choice of part queries. Thus,
even if a query might not be semantically correct, as long as it is useful for separating the object into distinct parts,
you should score it high.
- Bad queries would result in bad segmentation that are noisy or different parts are not correctly and/or clearly separated.

Your output should be a JSON in the following format:

```json
{
  "query_evaluations": {
    "query_0": {
      "score": 8,
      "reasoning": "This query effectively separates the object into functionally distinct parts. The boundaries are clean and consistent across different views."
    },
    "query_1": {
      "score": 3,
      "reasoning": "This query fails to distinguish important parts of the object, making it unsuitable for physical property assignment."
    },
    ...
  },
  "best_query": "query_1",
  "suggested_improvements": "Consider using more specific terms like 'ceramic pot' instead of just 'pot' to improve segmentation boundaries."
}
```
where `query_{i}` is the i-th query in the "all_queries" list.

Be detailed in your reasoning and make concrete suggestions for improvements.
\end{lstlisting}
}
\end{tcolorbox}
\caption{\textbf{VLM Critic System Prompt.} System instruction for evaluating segmentation quality and suggesting improvements.}
\label{fig:vlm_seg_critic_prompt}
\end{figure} 

%% file: prompt_incontext_physics.tex
\clearpage
\begin{lstlisting}[
    style=mystyle,
    caption={In-context Physics Examples},
    label={lst:incontext_examples}
]
{
    "tree": {
        "class_name_for_example": "ficus tree",
        "special_notes": "",
        "example_material_dict": {
            "pot": {"density": 400, "E": 2e8, "nu": 0.4, "material_id": get_material_id("rigid")},
            "trunk": {"density": 400, "E": 2e6, "nu": 0.4, "material_id": get_material_id("elastic")},
            "leaves": {"density": 200, "E": 2e4, "nu": 0.4, "material_id": get_material_id("elastic")}
        },
        "example_explanation": textwrap.dedent("""
            For this, we assume that the pot is stationary, while the trunk and leaves are made of "elastic", which will make
            them sway in the wind. The stiffness (Young's Modulus) of the trunk is much higher than that of the leaves.
        """),
        "example_all_queries": [["leaves", "trunk", "pot"], ["green", "orange", "reddish-brown"]],
        "tips": [
            "In a scene, typically there's a stationary part that will serve to fix the object to the ground. Usually, it's the pot, or some base of the tree. You must set the material_id of the stationary part to 6. If there's no stationary part, then never mind.",
            "The higher the `E` is, the stiffer the object is. E.g., so tree would sway less in the wind.",
        ]
    },
    "flowers": {
        "class_name_for_example": "flowers in a vase",
        "special_notes": "",
        "example_material_dict": {
            "vase": {"density": 500, "E": 1e6, "nu": 0.3, "material_id": get_material_id("rigid")},
            "flowers": {"density": 100, "E": 1e4, "nu": 0.4, "material_id": get_material_id("elastic")}
        },
        "example_explanation": textwrap.dedent("""
            Here, the vase is designated as stationary (material_id=6), indicating it should not move or sway.
            The flowers are set to a more pliable or flexible material (like "elastic" = 0), so that they can sway
            if there's wind or slight motion. The stiffness (Young's Modulus) of the vase is much higher than that
            of the flowers, making the vase rigid and the flowers more flexible.
        """),
        "example_all_queries": [["vase", "flowers"], ["ceramic base", "petals"], ["blue vase", "pink flower"]],
        "tips": [
            "In a typical flower arrangement, the vase (or base) is stationary, so give that part material_id=6 if present.",
            "The higher the `E`, the stiffer the part. So the vase should have a higher E range than the flowers.",
        ]
    },
    "shrub": {
        "class_name_for_example": "typical three-part shrub",
        "special_notes": textwrap.dedent("""
        Dataset note: Shrubs in our dataset stand by themselves---there is no planter or base.
            You should therefore return only the shrub's structural parts and none of them are stationary.
        """),
        "example_material_dict": {
            "stems":    { "density": 300, "E": 1e5, "nu": 0.35, "material_id": get_material_id("elastic") },
            "twigs":    { "density": 250, "E": 6e4, "nu": 0.38, "material_id": get_material_id("elastic") },
            "foliage":  { "density": 150, "E": 2e4, "nu": 0.40, "material_id": get_material_id("elastic") }
        },
        "example_explanation": textwrap.dedent("""
            Return *ranges* instead of single values and accompany them with reasoning, Pythonic
            constraints, and alternative query lists.
        """),
        "example_all_queries": [
            ["stems", "twigs", "foliage"],
            ["woody stems", "thin branches", "leaves"],
            ["brown sticks", "small branches", "green leaves"]
        ],
        "tips": [
            "Provide exactly the parts visible (usually stems/twigs + foliage).",
            "1e4 <= E <= 1e6.",
            "Stems should be stiffest > twigs > foliage.",
            "No part uses material_id 6 because nothing is fixed to the ground.",
        ]
    },
    "grass": {
        "class_name_for_example": "",
        "special_notes": textwrap.dedent("""
            **Dataset note:** Grass patches are usually isolated; occasionally a visible soil patch is
            underneath. Include a "soil" part only if it is visible.
        """),
        "example_material_dict": {
            "blades": { "density": 80, "E": 1e4, "nu": 0.45, "material_id": get_material_id("elastic") }
        },
        "example_explanation": textwrap.dedent("""
            Example A (typical isolated grass---no stationary part):
            ```json
            {
                "blades": { "density": 80, "E": 1e4, "nu": 0.45, "material_id": get_material_id("elastic") }
            }
            ```

            Example B (grass with visible soil):
            ```json
            {
                "soil":   { "density": 1200, "E": 5e5, "nu": 0.30, "material_id": get_material_id("rigid") },
                "blades": { "density":  80,  "E": 1e4, "nu": 0.45, "material_id": get_material_id("elastic") }
            }
            ```
            Return *ranges*, reasoning, constraints, and alternative query lists.
        """),
        "example_all_queries": [
          ["blades"],
          ["grass"],
          ["green stalks"]
        ],
        "tips": [
            "Segment only the visible parts (sometimes just \"blades\").",
            "If *no* soil visible:\nall_queries: [[\"blades\"],[\"grass\"],[\"green stalks\"]]",
            "If soil *is* visible:\nall_queries: [[\"soil\", \"blades\"],[\"dirt\", \"grass\"],[\"brown base\", \"green grass\"]]",
            "1e4 <= E <= 1e6.",
            "If soil present -> give it material_id 6 and ensure E_soil > E_blades.",
            "If soil absent -> no stationary part; material_id 6 should not appear.",
        ]
    },
    "rubber_ducks_and_toys": {
        "class_name_for_example": "",
        "special_notes": textwrap.dedent("""
            IMPORTANT: For rubber ducks and toys, we want to treat the entire object as a single part. Do not attempt to
            segment it into multiple parts. The object should be treated as a single, bouncy rubber-like object.
        """),
        "example_material_dict": {
            "toy": {"density": [80, 150], "E": [3e4, 5e4], "nu": [0.4, 0.45], "material_id": get_material_id("elastic")}
        },
        "example_explanation": "",
        "example_all_queries": [["toy"], ["rubber toy"], ["yellow duck"], ["plastic toy"]],
        "tips": [
            "Always use material_id=0 (jelly) for bouncy rubber-like behavior",
            "Keep E relatively low (around 1e3) for good bounce",
            "Density should be in the range of typical rubber/plastic toys",
            "Poisson's ratio should be around 0.35 for rubber-like behavior",
            "Make sure all queries in all_queries list are single-part queries"
        ]
    },
    "sport_balls": {
        "class_name_for_example": "",
        "special_notes": textwrap.dedent("""
            IMPORTANT: For sport balls, we want to treat the entire ball as a single part. Do not attempt to
            segment it into multiple parts (like surface patterns or seams). The ball should be treated as a single,
            bouncy object.
        """),
        "example_material_dict": {
            "ball": {"density": [80, 150], "E": [3e4, 5e4], "nu": [0.4, 0.45], "material_id": get_material_id("elastic")}
        },
        "example_explanation": "",
        "example_all_queries": [["ball"], ["sport ball"], ["basketball"], ["round ball"]],
        "tips": [
            "Always use material_id=0 (jelly) for bouncy behavior",
            "Keep E relatively low (around 1e3) for good bounce",
            "Density should be in the range of typical sport balls",
            "Poisson's ratio should be around 0.35 for rubber-like behavior",
            "Make sure all queries in all_queries list are single-part queries"
        ]
    },
    "soda_cans": {
        "class_name_for_example": "",
        "special_notes": textwrap.dedent("""
            IMPORTANT: For soda cans, we want to treat the entire can as a single part. Do not attempt to
            segment it into multiple parts (like the top, body, or label). The can should be treated as a single,
            rigid metal object.
        """),
        "example_material_dict": {
            "can": {"density": [2600, 2800], "E": [5e10, 8e10], "nu": [0.25, 0.35], "material_id": get_material_id("metal")}
        },
        "example_explanation": "",
        "example_all_queries": [["can"], ["soda can"], ["aluminum can"], ["metal can"]],
        "tips": [
            "Always use material_id=1 (metal) for rigid metal behavior",
            "Keep E relatively high (around 1e8) for metal stiffness",
            "Density should be in the range of typical aluminum (around 2700 kg/m^3)",
            "Poisson's ratio should be around 0.3 for metal behavior",
            "Make sure all queries in all_queries list are single-part queries"
        ]
    },
    "metal_crates": {
        "class_name_for_example": "",
        "special_notes": textwrap.dedent("""
            IMPORTANT: For metal crates, we want to treat the entire crate as a single part. Do not attempt to
            segment it into multiple parts (like the sides, top, or bottom). The crate should be treated as a single,
            rigid metal object.
        """),
        "example_material_dict": {
            "crate": {"density": [2500, 2900], "E": [8e7, 1.2e8], "nu": [0.25, 0.35], "material_id": get_material_id("metal")}
        },
        "example_explanation": "",
        "example_all_queries": [["crate"], ["metal crate"], ["metal box"], ["steel crate"]],
        "tips": [
            "Always use material_id=1 (metal) for rigid metal behavior",
            "Keep E relatively high (around 1e8) for metal stiffness",
            "Density should be in the range of typical metal (around 2700 kg/m^3)",
            "Poisson's ratio should be around 0.3 for metal behavior",
            "Make sure all queries in all_queries list are single-part queries"
        ]
    },
    "sand": {
        "class_name_for_example": "",
        "special_notes": textwrap.dedent("""
            IMPORTANT: For sand objects, we want to treat the entire object as a single part. Do not attempt to
            segment it into multiple parts. The sand should be treated as a single, granular material.
        """),
        "example_material_dict": {
            "sand": {"density": [1800, 2200], "E": [4e7, 6e7], "nu": [0.25, 0.35], "material_id": get_material_id("sand")}
        },
        "example_explanation": "",
        "example_all_queries": [["sand"], ["sand pile"], ["sand mound"], ["granular material"]],
        "tips": [
            "Always use material_id=2 (sand) for granular behavior",
            "Keep E relatively high (around 5e7) for sand stiffness",
            "Density should be in the range of typical sand (around 2000 kg/m^3)",
            "Poisson's ratio should be around 0.3 for sand behavior",
            "Make sure all queries in all_queries list are single-part queries"
        ]
    },
    "jello_block": {
        "class_name_for_example": "",
        "special_notes": textwrap.dedent("""
            IMPORTANT: For jello blocks, we want to treat the entire object as a single part. Do not attempt to
            segment it into multiple parts. The jello block should be treated as a single, soft, bouncy object.
        """),
        "example_material_dict": {
            "jello": {"density": [40, 60], "E": [800, 1200], "nu": [0.25, 0.35], "material_id": get_material_id("elastic")}
        },
        "example_explanation": "",
        "example_all_queries": [["jello"], ["jello block"], ["gelatin"], ["bouncy block"]],
        "tips": [
            "Always use material_id=0 (jelly) for soft, bouncy behavior",
            "Keep E relatively low (around 1000) for good bounce and jiggle",
            "Density should be in the range of typical jello (around 50 kg/m^3)",
            "Poisson's ratio should be around 0.3 for jello-like behavior",
            "Make sure all queries in all_queries list are single-part queries"
        ]
    },
    "snow_and_mud": {
        "class_name_for_example": "",
        "special_notes": textwrap.dedent("""
            IMPORTANT: For combined snow & mud objects, we treat the entire mixture as a single deformable part.  Do **not**
            attempt to split it into separate snow and mud regions---the simulation will use one MPM material.
        """),
        "example_material_dict": {
            "snow_and_mud": {"density": [2000, 3000], "E": [8e4, 1.2e5], "nu": [0.15, 0.25], "material_id": get_material_id("snow")}
        },
        "example_explanation": "",
        "example_all_queries": [["snow and mud"], ["slush"], ["muddy snow"], ["wet snow"]],
        "tips": [
            "Always set material_id = 5 (snow) so the simulator uses the appropriate elasto-plastic snow model.",
            "Keep E around 1e5 (the config value) to match the intended softness.",
            "Density is markedly higher than fluffy snow because of the mud/water content---use roughly 2-3 g/cm^3 (2000-3000 kg/m^3).",
            "Make sure every list in `all_queries` contains **one** phrase because this is a single-part object."
        ]
    },
}
\end{lstlisting}

%% file: human_prior.tex
\section{The Effects of Human Prior on \dataset} \label{sec:appendix_nerf2physic_ours_data}

\dataset is labeled via VLMs using in-conext physics examples manually tuned by humans. A condensed version of these in-context examples is provided in Fig.~\ref{fig:incontext_physics_toy_examples} and the full prompt in Listing~\ref{lst:incontext_examples}. These examples align the VLM's physical understanding with human's. In our ablation result, we found that removing these examples significantly results as shown in Tab.~\ref{tab:vlm-exec-2col}. 

The main differences between \dataset labeling and NeRF2Physics are

\begin{enumerate}
    \item We use VLM to propose object-dependent segmentation while NeRF2Physics using LLM is essentially blind. Specifically, ur VLM actor proposes segmentation queries based on a set of mutli-view images of the object as shown in Fig.~\ref{fig:vlm_actor}.
\item We use semantic proposals (e.g., "pot", "trunk") instead of material proposals (e.g., "leather", "stone") like NeRF2Physics did. Computing similarity directly between material name and CLIP features yields inaccurate and noisy segmentation as shown in Fig.~\ref{fig:clip_seg}. This also limits the generality of the NeRF2Physics since one material type (e.g., ``elastic") can only have a fixed set of parameters in a scene. In contrast, \method enables spatially-varying parameter specification: the leaves and the trunk of a tree while both belonging to the same ``elastic" class can have vastly different young modulus, Poisson ratio and density as shown in Fig.~\ref{fig:ours_pred}.
\item We proposes multiple candidates (e.g., "pot,leaves" vs "base,folliage") and use a VLM critic to select the best based on CLIP-based segmentation while NeRF2Physics does not have any selection mechanism. Figure \ref{fig:clip_seg} show the dramatic segmentation quality across different queries, highlighting the need for multiple candidates and selection.
\item We also provide manually tuned in-context physics parameter examples.
\end{enumerate}

These crucial differences contribute to much higher quality dataset labeling as shown in Tab.~\ref{tab:vlm-exec-2col}.

\begin{table}[b]
\centering
\small
\setlength{\tabcolsep}{8pt}
\caption{\textbf{\dataset Ablation.} The effect of in-context physics examples on data quality. We include the executionability rate, which computes the fraction of times that a physic simulation can be successfully run without numerical explosion, and the realism score judged by Gemini.}
\vspace{4pt}
\begin{tabular}{lcc}
\toprule
\textbf{Method} & \textbf{Exec.\ Rate $\uparrow$} & \textbf{VLM Score $\uparrow$} \\
\midrule
\textbf{W/ In-context Examples (Ours)}            & 100.0\% & \meanstd{4.83}{0.09} \\
W/o In-context Examples                     & 62.5\%  & \meanstd{1.34}{0.30} \\
NeRF2Physics~\citep{zhai2024physical} & 45.0\%  & \meanstd{1.09}{0.28} \\
\bottomrule
\end{tabular}
\label{tab:vlm-exec-2col}
\vspace{-4pt}
\end{table}

%% file: vlm_judge.tex
\section{VLM As a Physics Judge} \label{sec:vlm_judge}
We utilize a VLM to evaluate the realism of different candidate videos. The videos are scored on the scale 1-5, and an optional reference video and the prompt describing the video (e.g,. ``tree swaying in the wind") is provided. We also use Cotracker \citep{karaev2024cotracker} to annotate the videos with motion traces. The system prompt is provided in Fig.~\ref{fig:prompt_vlm_eval}.

\input{prompt_vlm_eval}

%% file: prompt_vlm_eval.tex
\begin{figure}
    \centering
\begin{tcolorbox}[top=2pt,bottom=2pt,width=\linewidth,boxrule=1pt]{
\begin{lstlisting}[style=promptstyle]
You are a physics-realism judge for animation videos.

You will be shown several candidate animations of the SAME 3D object responding to the SAME textual prompt that describes its intended physical motion.

Your tasks:
1. Carefully watch each candidate animation.
2. Describe what's going on in the animation.
3. Evaluate how physically realistic the motion looks (0-5 scale).
4. Identify concrete pros / cons affecting the score (e.g. energy conservation errors, temporal jitter, incorrect response to gravity, static etc.).
5. Suggest specific improvements.
6. Pick the overall best candidate.

Please output ONLY valid JSON with the following schema:
{
  "candidate_evaluations": {
    "candidate_0": {"description": str, "score": float, "pros": str, "cons": str, "suggested_improvements": str},
    "candidate_1": { ... },
    "candidate_2": { ... }
  },
  "best_candidate": "candidate_i",   // the key of the best candidate
  "general_comments": str              // any overall remarks (optional)
}

NOTE: ignore missing videos. Still return score for `candidate_{idx}` that are present.

NOTE: to make your job easier, we have also annotated the video with the Co-Tracker. Cotracker is a motion tracker algorithm to highlight the moving parts in the videos. 
Pay close attention to the motion traces annotated in the videos to gain information on how the object is moving.
Note that for objects that barely move, there will still be dots in the Co-Tracker video, but the motion
(lines) will be very short or non-existent, indicating that the points are not moving.

Cotracker can sometimes produce noisy traces so only use it as a reference, and consider the motion of the object as a whole, and other visual cues.
\end{lstlisting}
}
\end{tcolorbox}
\caption{\textbf{VLM Evaluator's System Prompt.} }
\label{fig:prompt_vlm_eval}
\end{figure} 

%% file: model_arch.tex
\section{Model architecture} \label{sec:model-arch}

\subsection{Overview}
We employ a 3D UNet-based architecture for both discrete material segmentation and continuous material parameter regression. The architecture consists of two main components: (1) a feature projector for dimensionality reduction, and (2) a 3D UNet backbone for spatial processing.

\subsection{Feature Projector}
The feature projector is used when the input feature dimension differs from the conditioning dimension:
\begin{itemize}
    \item \textbf{Input features}: The model supports three input modalities:
    \begin{itemize}
        \item RGB features: $\mathbf{F} \in \mathbb{R}^{N \times 3 \times D \times H \times W}$
        \item CLIP features: $\mathbf{F} \in \mathbb{R}^{N \times 768 \times D \times H \times W}$  
        \item Occupancy features: $\mathbf{F} \in \mathbb{R}^{N \times 1 \times D \times H \times W}$
    \end{itemize}
    \item \textbf{Projection}: Features are projected to a unified conditioning dimension of 32 channels using a feature projector with hidden dimension of 128 (when input channels $>$ 32). The projector consists of three layers of Conv3D, GroupNorm and SiLU activation.
\end{itemize}

\subsection{3D UNet Architecture}

We employ a U-Net architecture \citep{dhariwal2021diffusion, ronneberger2015u} operating on 3D feature grids of shape $\mathbb{R}^{N \times 32 \times 64 \times 64 \times 64}$. The network follows a standard encoder-decoder structure with skip connections, using a base channel dimension of 64 and channel multipliers of [1, 1, 2, 4] across four resolution levels.

The encoder begins with a 3D convolution that projects the 32-dimensional input features to 64 channels. The encoder then processes features through four resolution levels, each containing three residual blocks. The first two levels maintain 64 channels while progressively reducing spatial dimensions from $64^3$ to $32^3$. The subsequent levels double the channel count at each downsampling step, reaching 128 channels at $16^3$ resolution and 256 channels at $8^3$ resolution. Downsampling between levels is performed using strided 3D convolutions with stride 2.

At the bottleneck, the network processes the lowest resolution features through a sequence of residual block, attention block, and another residual block, all operating at $8^3$ spatial resolution with 256 channels. Note that in our implementation, attention blocks are disabled by setting attention resolutions to empty.

The decoder symmetrically reverses the encoder path, utilizing skip connections from corresponding encoder levels. Upsampling is achieved through nearest-neighbor interpolation with a scale factor of 2, followed by 3D convolution. Each decoder level matches the channel dimensions and number of residual blocks of its corresponding encoder level.

Each residual block follows the formulation $\text{ResBlock}(x) = x + f(x)$, where $f$ consists of layer normalization, LeakyReLU activation with negative slope 0.02, 3D convolution with kernel size 3, another layer normalization and activation, dropout, and a final zero-initialized 3D convolution. When input and output channels differ, the skip connection employs a $1 \times 1 \times 1$ convolution for channel matching.

The final output layer applies layer normalization, LeakyReLU activation, and a 3D convolution that projects to either 8 channels for discrete material classification or 3 channels for continuous material parameter regression.

%% file: appendix_result.tex
\section{Additional Results} \label{sec:more_res}

\begin{figure}[!t]
        \vspace{-1cm}
        \centering
        \includegraphics[width=1.\textwidth, trim={0 0cm 0 0cm}, clip]{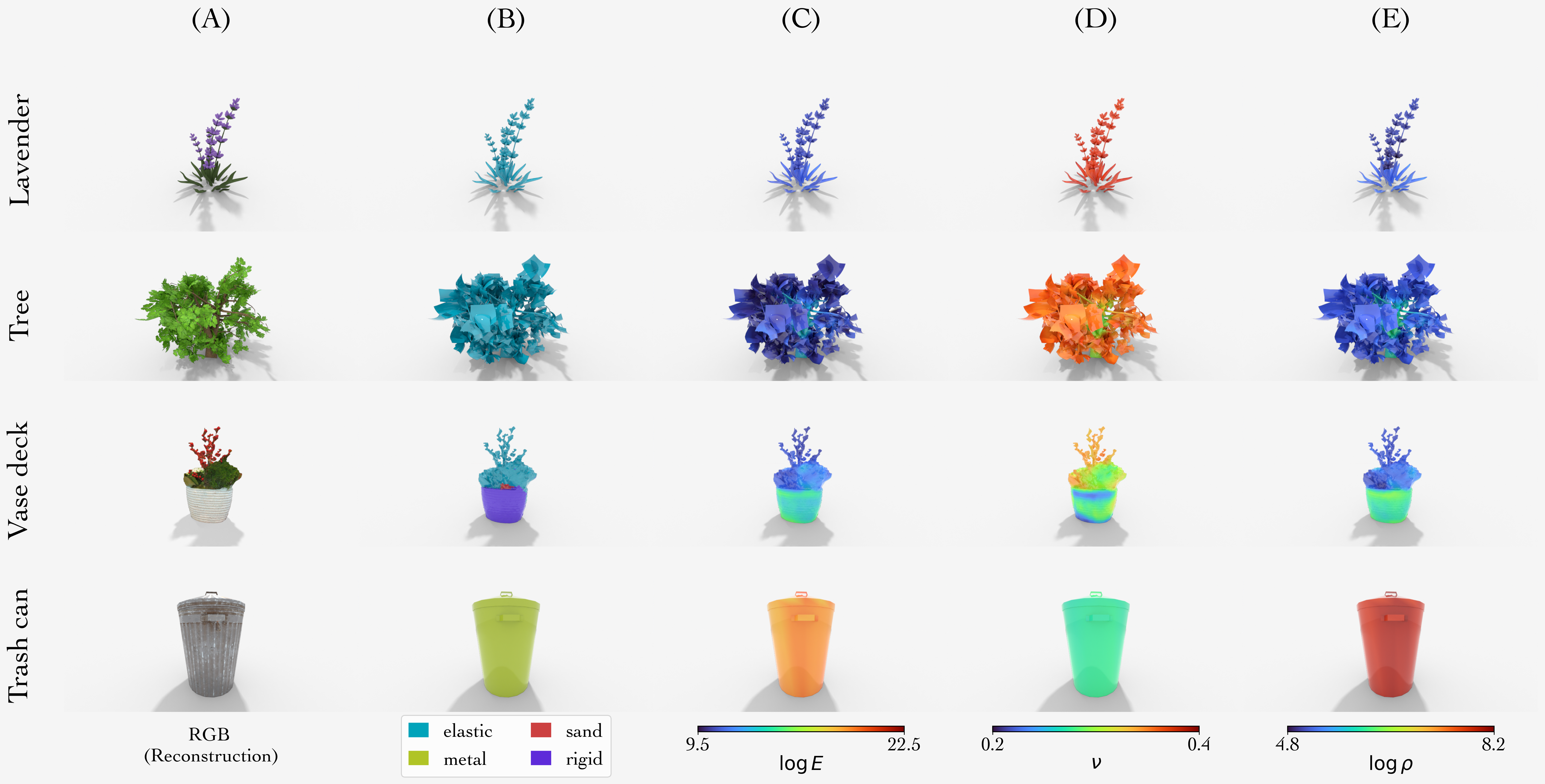}
        \includegraphics[width=1.\textwidth, trim={0 0cm 0 0cm}, clip]{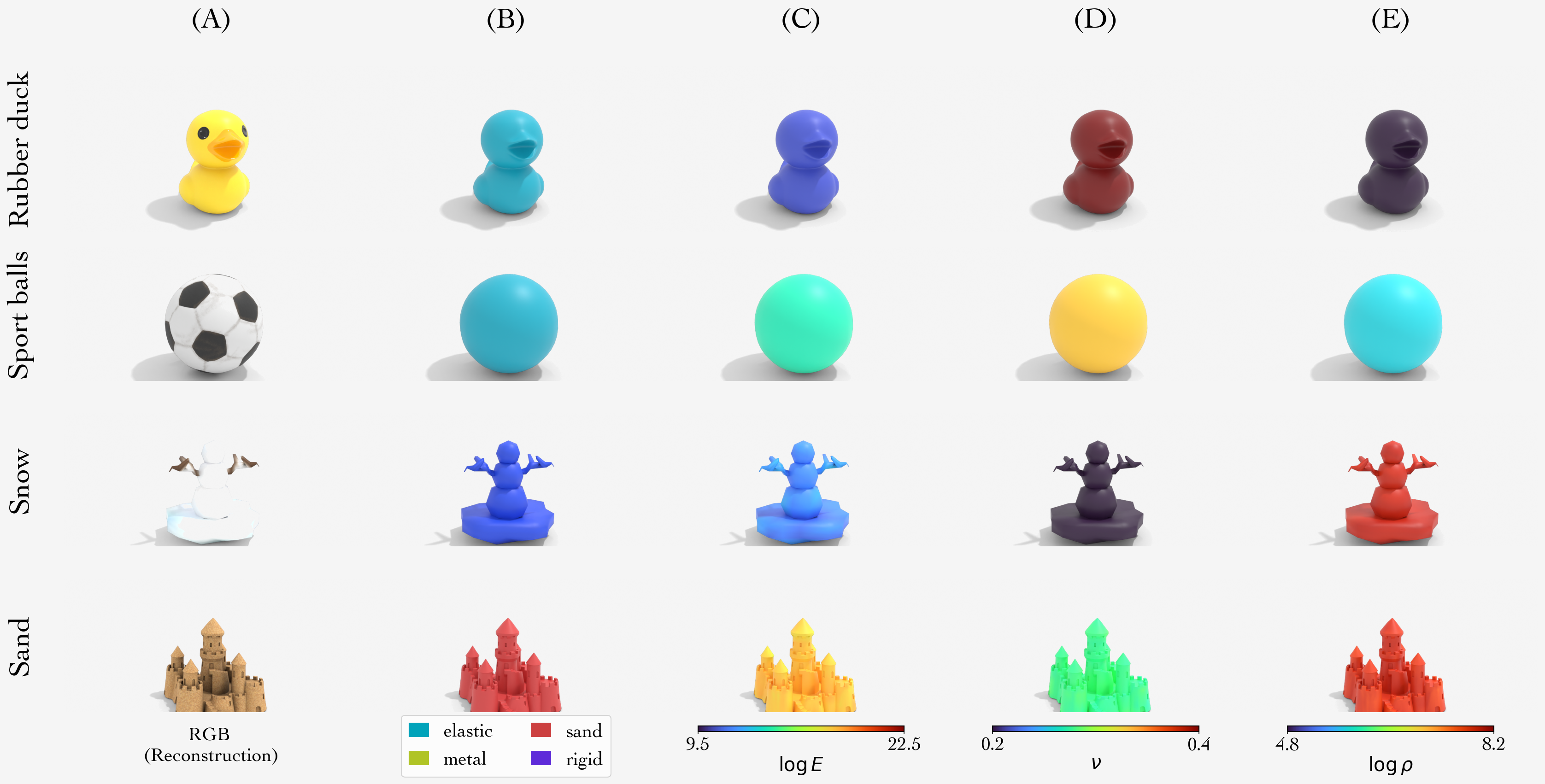}
       \caption{\textbf{\method Prediction Visualization.} \method simultaneously recovers discrete material class, continuous Young's modulus (E), Poisson's ratio ($\nu$), and mass density ($\rho$) with a high degree of accuracy. For example, the model correctly labels foliage as elastic and the metal can as rigid, while recovering realistic stiffness and density gradients within each object.}
        \label{fig:ours_pred}
\end{figure}

\begin{figure}[!t]
        \centering
        \includegraphics[width=1.\textwidth, trim={0 0cm 0 0cm}, clip]{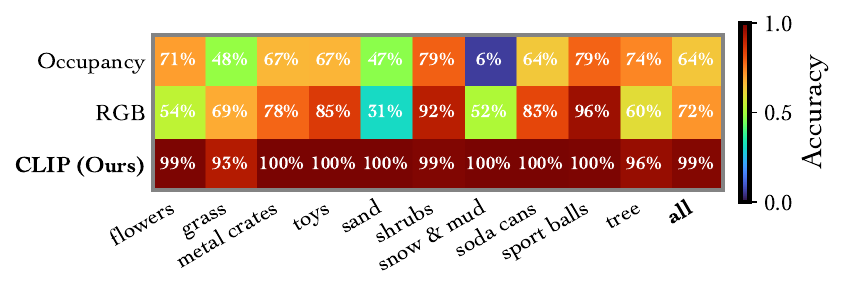}
       \caption{\textbf{\method Ablation's Per-class Accuracy on synthetic scenes}. CLIP features generalizes in synthetic scenes, outperforming RGB and occupancy on all classes.}
        \label{fig:ablation_acc}
\end{figure}

\begin{figure}[!t]
\vspace{-2cm}
        \centering
         \includegraphics[width=1\textwidth, trim={0 0 0 0}]{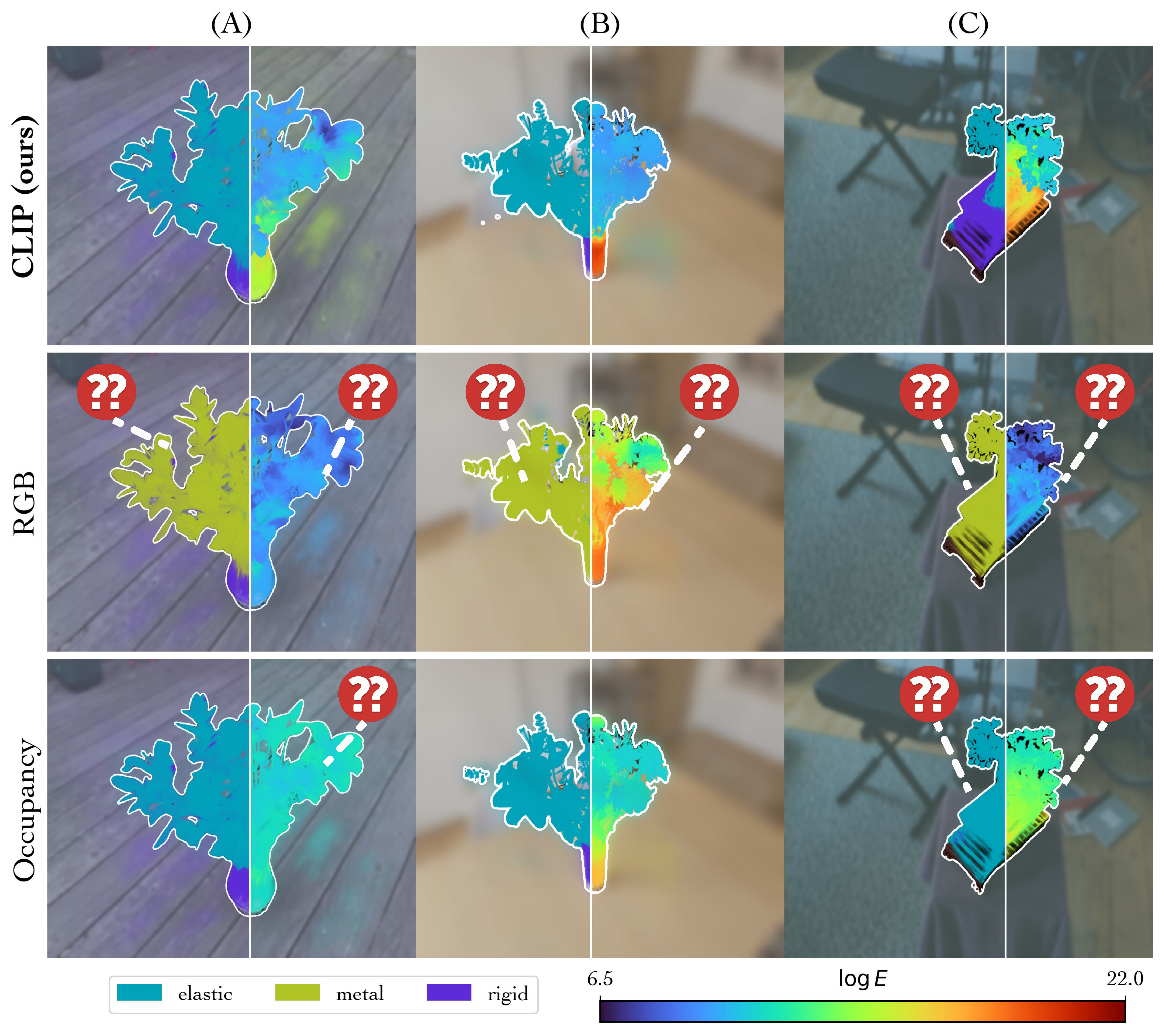}
        \includegraphics[width=1\textwidth, trim={0 0 0 0}]{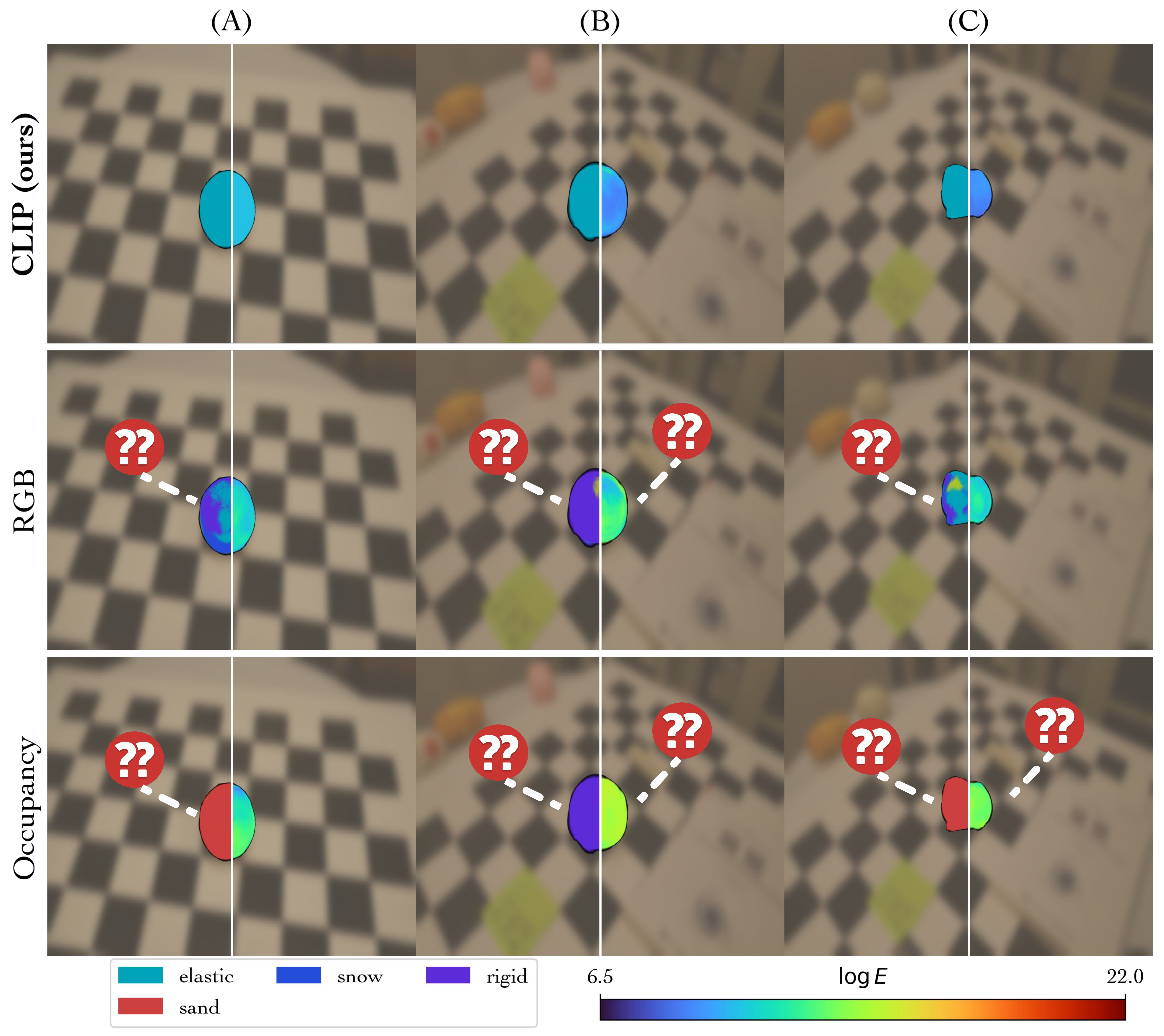}

    \caption{\textbf{\method's Feature Type Ablation on Real Scenes.} Replacing CLIP features with RGB or occupancy severely degrades the material prediction. Incorrect predictions such as leave mislaballed as metal or Young's modulus being uniform within an object are marked with question marks. This highlights the power of pretrained visual features in bridging the sim2real gap.}
    \label{fig:real_res_ablation}
\end{figure}

We visualize the physics predictions by our model in Fig.~\ref{fig:ours_pred}. Figure \ref{fig:ablation_acc} breaks down the material accuracy across semantic classes of \dataset between our \method CLIP versus two ablated versions using RGB and occupancy input features. Figure \ref{fig:real_res_ablation} qualitatively compare the ablated methods on the real-world scenes.

%% file: neurips_2024.bbl
\begin{thebibliography}{45}
\providecommand{\natexlab}[1]{#1}
\providecommand{\url}[1]{\texttt{#1}}
\expandafter\ifx\csname urlstyle\endcsname\relax
  \providecommand{\doi}[1]{doi: #1}\else
  \providecommand{\doi}{doi: \begingroup \urlstyle{rm}\Url}\fi

\bibitem[Abou-Chakra et~al.(2024)Abou-Chakra, Rana, Dayoub, and Suenderhauf]{abou-chakra2024physically}
Jad Abou-Chakra, Krishan Rana, Feras Dayoub, and Niko Suenderhauf.
\newblock Physically embodied gaussian splatting: A realtime correctable world model for robotics.
\newblock In \emph{8th Annual Conference on Robot Learning}, 2024.
\newblock URL \url{https://openreview.net/forum?id=AEq0onGrN2}.

\bibitem[Bear et~al.(2021)Bear, Wang, Mrowca, Binder, Tung, Pramod, Holdaway, Tao, Smith, Sun, et~al.]{bear2021physion}
Daniel~M Bear, Elias Wang, Damian Mrowca, Felix~J Binder, Hsiao-Yu~Fish Tung, RT~Pramod, Cameron Holdaway, Sirui Tao, Kevin Smith, Fan-Yun Sun, et~al.
\newblock Physion: Evaluating physical prediction from vision in humans and machines.
\newblock \emph{arXiv preprint arXiv:2106.08261}, 2021.

\bibitem[Bell et~al.(2015)Bell, Upchurch, Snavely, and Bala]{bell2015material}
Sean Bell, Paul Upchurch, Noah Snavely, and Kavita Bala.
\newblock Material recognition in the wild with the materials in context database.
\newblock In \emph{Proceedings of the IEEE conference on computer vision and pattern recognition}, pages 3479--3487, 2015.

\bibitem[Cao et~al.(2025)Cao, Chen, Pan, and Liu]{cao2025physx}
Ziang Cao, Zhaoxi Chen, Liang Pan, and Ziwei Liu.
\newblock Physx: Physical-grounded 3d asset generation.
\newblock \emph{arXiv preprint arXiv:2507.12465}, 2025.

\bibitem[Chen et~al.(2025{\natexlab{a}})Chen, Jiang, Liu, Gupta, Li, Zhao, and Wang]{chen2025physgen3d}
Boyuan Chen, Hanxiao Jiang, Shaowei Liu, Saurabh Gupta, Yunzhu Li, Hao Zhao, and Shenlong Wang.
\newblock Physgen3d: Crafting a miniature interactive world from a single image.
\newblock \emph{arXiv preprint arXiv:2503.20746}, 2025{\natexlab{a}}.

\bibitem[Chen et~al.(2025{\natexlab{b}})Chen, Dou, Wang, Huang, Chen, Feng, Gu, and Liu]{chen2025vid2sim}
Chuhao Chen, Zhiyang Dou, Chen Wang, Yiming Huang, Anjun Chen, Qiao Feng, Jiatao Gu, and Lingjie Liu.
\newblock Vid2sim: Generalizable, video-based reconstruction of appearance, geometry and physics for mesh-free simulation.
\newblock \emph{IEEE Conference on Computer Vision and Pattern Recognition (CVPR)}, 2025{\natexlab{b}}.

\bibitem[Chen et~al.(2024)Chen, Zhang, Zhang, Zhou, Kim, Liu, Li, Zhang, Zhao, Wang, et~al.]{chen2024unireal}
Xi~Chen, Zhifei Zhang, He~Zhang, Yuqian Zhou, Soo~Ye Kim, Qing Liu, Yijun Li, Jianming Zhang, Nanxuan Zhao, Yilin Wang, et~al.
\newblock Unireal: Universal image generation and editing via learning real-world dynamics.
\newblock \emph{arXiv preprint arXiv:2412.07774}, 2024.

\bibitem[Deitke et~al.(2022)Deitke, Schwenk, Salvador, Weihs, Michel, VanderBilt, Schmidt, Ehsani, Kembhavi, and Farhadi]{deitke2022objaverseuniverseannotated3d}
Matt Deitke, Dustin Schwenk, Jordi Salvador, Luca Weihs, Oscar Michel, Eli VanderBilt, Ludwig Schmidt, Kiana Ehsani, Aniruddha Kembhavi, and Ali Farhadi.
\newblock Objaverse: A universe of annotated 3d objects, 2022.
\newblock URL \url{https://arxiv.org/abs/2212.08051}.

\bibitem[Dhariwal and Nichol(2021)]{dhariwal2021diffusion}
Prafulla Dhariwal and Alexander Nichol.
\newblock Diffusion models beat gans on image synthesis.
\newblock \emph{Advances in neural information processing systems}, 34:\penalty0 8780--8794, 2021.

\bibitem[Feng et~al.(2023)Feng, Shang, Li, Shao, Jiang, and Yang]{feng2023pienerf}
Yutao Feng, Yintong Shang, Xuan Li, Tianjia Shao, Chenfanfu Jiang, and Yin Yang.
\newblock Pie-nerf: Physics-based interactive elastodynamics with nerf, 2023.

\bibitem[Fischer et~al.(2024)Fischer, Georgiev, Groueix, Kim, Ritschel, and Deschaintre]{fischer2024sama}
Michael Fischer, Iliyan Georgiev, Thibault Groueix, Vladimir~G Kim, Tobias Ritschel, and Valentin Deschaintre.
\newblock Sama: Material-aware 3d selection and segmentation.
\newblock \emph{arXiv preprint arXiv:2411.19322}, 2024.

\bibitem[Guo et~al.(2024)Guo, Wang, Ma, Zhang, Owens, Gan, Tenenbaum, He, and Matusik]{guo2024physcomp}
Minghao Guo, Bohan Wang, Pingchuan Ma, Tianyuan Zhang, Crystal~Elaine Owens, Chuang Gan, Joshua~B. Tenenbaum, Kaiming He, and Wojciech Matusik.
\newblock Physically compatible 3d object modeling from a single image.
\newblock \emph{arXiv preprint arXiv:2405.20510}, 2024.

\bibitem[Hsu et~al.(2024)Hsu, Lin, Zhai, Xia, and Wang]{hsu2024autovfx}
Hao-Yu Hsu, Zhi-Hao Lin, Albert Zhai, Hongchi Xia, and Shenlong Wang.
\newblock Autovfx: Physically realistic video editing from natural language instructions.
\newblock \emph{arXiv preprint arXiv:2411.02394}, 2024.

\bibitem[Huang et~al.(2024)Huang, Zeng, Li, Zuo, and Lau]{huang2024dreamphysics}
Tianyu Huang, Yihan Zeng, Hui Li, Wangmeng Zuo, and Rynson~WH Lau.
\newblock Dreamphysics: Learning physical properties of dynamic 3d gaussians with video diffusion priors.
\newblock \emph{arXiv preprint arXiv:2406.01476}, 2024.

\bibitem[Jatavallabhula et~al.(2021)Jatavallabhula, Macklin, Golemo, Voleti, Petrini, Weiss, Considine, Parent-Levesque, Xie, Erleben, Paull, Shkurti, Nowrouzezahrai, and Fidler]{gradsim}
Krishna~Murthy Jatavallabhula, Miles Macklin, Florian Golemo, Vikram Voleti, Linda Petrini, Martin Weiss, Breandan Considine, Jerome Parent-Levesque, Kevin Xie, Kenny Erleben, Liam Paull, Florian Shkurti, Derek Nowrouzezahrai, and Sanja Fidler.
\newblock gradsim: Differentiable simulation for system identification and visuomotor control.
\newblock \emph{International Conference on Learning Representations (ICLR)}, 2021.
\newblock URL \url{https://openreview.net/forum?id=c_E8kFWfhp0}.

\bibitem[Jiang et~al.(2025)Jiang, Hsu, Zhang, Yu, Wang, and Li]{jiang2025phystwin}
Hanxiao Jiang, Hao-Yu Hsu, Kaifeng Zhang, Hsin-Ni Yu, Shenlong Wang, and Yunzhu Li.
\newblock Phystwin: Physics-informed reconstruction and simulation of deformable objects from videos.
\newblock \emph{arXiv preprint arXiv:2503.17973}, 2025.

\bibitem[Karaev et~al.(2024)Karaev, Rocco, Graham, Neverova, Vedaldi, and Rupprecht]{karaev2024cotracker}
Nikita Karaev, Ignacio Rocco, Benjamin Graham, Natalia Neverova, Andrea Vedaldi, and Christian Rupprecht.
\newblock Cotracker: It is better to track together.
\newblock In \emph{European Conference on Computer Vision}, pages 18--35. Springer, 2024.

\bibitem[Kerbl et~al.(2023)Kerbl, Kopanas, Leimk{\"u}hler, and Drettakis]{kerbl20233d}
Bernhard Kerbl, Georgios Kopanas, Thomas Leimk{\"u}hler, and George Drettakis.
\newblock 3d gaussian splatting for real-time radiance field rendering.
\newblock \emph{ACM Trans. Graph.}, 42\penalty0 (4):\penalty0 139--1, 2023.

\bibitem[Kerr et~al.(2023)Kerr, Kim, Goldberg, Kanazawa, and Tancik]{kerr2023lerf}
Justin Kerr, Chung~Min Kim, Ken Goldberg, Angjoo Kanazawa, and Matthew Tancik.
\newblock Lerf: Language embedded radiance fields.
\newblock In \emph{Proceedings of the IEEE/CVF International Conference on Computer Vision}, pages 19729--19739, 2023.

\bibitem[Kingma(2014)]{kingma2014adam}
Diederik~P Kingma.
\newblock Adam: A method for stochastic optimization.
\newblock \emph{arXiv preprint arXiv:1412.6980}, 2014.

\bibitem[Kobayashi et~al.(2022)Kobayashi, Matsumoto, and Sitzmann]{kobayashi2022distilledfeaturefields}
Sosuke Kobayashi, Eiichi Matsumoto, and Vincent Sitzmann.
\newblock Decomposing nerf for editing via feature field distillation.
\newblock In \emph{Advances in Neural Information Processing Systems}, volume~35, 2022.
\newblock URL \url{https://arxiv.org/pdf/2205.15585.pdf}.

\bibitem[Le et~al.(2024)Le, Xie, Liang, Wang, Yang, Ma, Vedder, Krishna, Jayaraman, and Eaton]{le2024articulate}
Long Le, Jason Xie, William Liang, Hung-Ju Wang, Yue Yang, Yecheng~Jason Ma, Kyle Vedder, Arjun Krishna, Dinesh Jayaraman, and Eric Eaton.
\newblock Articulate-anything: Automatic modeling of articulated objects via a vision-language foundation model.
\newblock \emph{arXiv preprint arXiv:2410.13882}, 2024.

\bibitem[Li et~al.(2023)Li, Qiao, Chen, Jatavallabhula, Lin, Jiang, and Gan]{li2023pacnerf}
Xuan Li, Yi-Ling Qiao, Peter~Yichen Chen, Krishna~Murthy Jatavallabhula, Ming Lin, Chenfanfu Jiang, and Chuang Gan.
\newblock {PAC}-ne{RF}: Physics augmented continuum neural radiance fields for geometry-agnostic system identification.
\newblock In \emph{The Eleventh International Conference on Learning Representations}, 2023.
\newblock URL \url{https://openreview.net/forum?id=tVkrbkz42vc}.

\bibitem[Li et~al.(2024)Li, Tucker, Snavely, and Holynski]{li2024generative}
Zhengqi Li, Richard Tucker, Noah Snavely, and Aleksander Holynski.
\newblock Generative image dynamics.
\newblock In \emph{Proceedings of the IEEE/CVF Conference on Computer Vision and Pattern Recognition}, pages 24142--24153, 2024.

\bibitem[Li et~al.(2025)Li, Yu, Liu, Yang, Herrmann, Wetzstein, and Wu]{li2025wonderplay}
Zizhang Li, Hong-Xing Yu, Wei Liu, Yin Yang, Charles Herrmann, Gordon Wetzstein, and Jiajun Wu.
\newblock Wonderplay: Dynamic 3d scene generation from a single image and actions.
\newblock \emph{arXiv preprint arXiv:2505.18151}, 2025.

\bibitem[Lin et~al.(2025)Lin, Lin, Xu, and MU]{lin2025omniphysgs}
Yuchen Lin, Chenguo Lin, Jianjin Xu, and Yadong MU.
\newblock Omniphys{GS}: 3d constitutive gaussians for general physics-based dynamics generation.
\newblock In \emph{The Thirteenth International Conference on Learning Representations}, 2025.
\newblock URL \url{https://openreview.net/forum?id=9HZtP6I5lv}.

\bibitem[Ma et~al.(2023)Ma, Chen, Deng, Tenenbaum, Du, Gan, and Matusik]{ma2023learning}
Pingchuan Ma, Peter~Yichen Chen, Bolei Deng, Joshua~B Tenenbaum, Tao Du, Chuang Gan, and Wojciech Matusik.
\newblock Learning neural constitutive laws from motion observations for generalizable pde dynamics.
\newblock In \emph{International Conference on Machine Learning}, pages 23279--23300. PMLR, 2023.

\bibitem[Mildenhall et~al.(2021)Mildenhall, Srinivasan, Tancik, Barron, Ramamoorthi, and Ng]{mildenhall2021nerf}
Ben Mildenhall, Pratul~P Srinivasan, Matthew Tancik, Jonathan~T Barron, Ravi Ramamoorthi, and Ren Ng.
\newblock Nerf: Representing scenes as neural radiance fields for view synthesis.
\newblock \emph{Communications of the ACM}, 65\penalty0 (1):\penalty0 99--106, 2021.

\bibitem[Mittal et~al.(2025)Mittal, Zhuang, Lee, and Tulsiani]{mittal2025uniphy}
Himangi Mittal, Peiye Zhuang, Hsin-Ying Lee, and Shubham Tulsiani.
\newblock Uniphy: Learning a unified constitutive model for inverse physics simulation.
\newblock \emph{arXiv preprint arXiv:2505.16971}, 2025.

\bibitem[Parker-Holder et~al.(2024)Parker-Holder, Ball, Bruce, Dasagi, Holsheimer, Kaplanis, Moufarek, Scully, Shar, Shi, Spencer, Yung, Dennis, Kenjeyev, Long, Mnih, Chan, Gazeau, Li, Pardo, Wang, Zhang, Besse, Harley, Mitenkova, Wang, Clune, Hassabis, Hadsell, Bolton, Singh, and Rockt{\"a}schel]{parkerholder2024genie2}
Jack Parker-Holder, Philip Ball, Jake Bruce, Vibhavari Dasagi, Kristian Holsheimer, Christos Kaplanis, Alexandre Moufarek, Guy Scully, Jeremy Shar, Jimmy Shi, Stephen Spencer, Jessica Yung, Michael Dennis, Sultan Kenjeyev, Shangbang Long, Vlad Mnih, Harris Chan, Maxime Gazeau, Bonnie Li, Fabio Pardo, Luyu Wang, Lei Zhang, Frederic Besse, Tim Harley, Anna Mitenkova, Jane Wang, Jeff Clune, Demis Hassabis, Raia Hadsell, Adrian Bolton, Satinder Singh, and Tim Rockt{\"a}schel.
\newblock Genie 2: A large-scale foundation world model.
\newblock 2024.
\newblock URL \url{https://deepmind.google/discover/blog/genie-2-a-large-scale-foundation-world-model/}.

\bibitem[Pumarola et~al.(2020)Pumarola, Corona, Pons-Moll, and Moreno-Noguer]{pumarola2020d}
Albert Pumarola, Enric Corona, Gerard Pons-Moll, and Francesc Moreno-Noguer.
\newblock {D-NeRF: Neural Radiance Fields for Dynamic Scenes}.
\newblock In \emph{Proceedings of the IEEE/CVF Conference on Computer Vision and Pattern Recognition}, 2020.

\bibitem[Qiu et~al.(2024)Qiu, Yang, Zeng, and Wang]{qiu2024feature}
Ri-Zhao Qiu, Ge~Yang, Weijia Zeng, and Xiaolong Wang.
\newblock Feature splatting: Language-driven physics-based scene synthesis and editing.
\newblock \emph{arXiv preprint arXiv:2404.01223}, 2024.

\bibitem[Radford et~al.(2021)Radford, Kim, Hallacy, Ramesh, Goh, Agarwal, Sastry, Askell, Mishkin, Clark, et~al.]{radford2021learning}
Alec Radford, Jong~Wook Kim, Chris Hallacy, Aditya Ramesh, Gabriel Goh, Sandhini Agarwal, Girish Sastry, Amanda Askell, Pamela Mishkin, Jack Clark, et~al.
\newblock Learning transferable visual models from natural language supervision.
\newblock In \emph{International conference on machine learning}, pages 8748--8763. PmLR, 2021.

\bibitem[Ronneberger et~al.(2015)Ronneberger, Fischer, and Brox]{ronneberger2015u}
Olaf Ronneberger, Philipp Fischer, and Thomas Brox.
\newblock U-net: Convolutional networks for biomedical image segmentation.
\newblock In \emph{International Conference on Medical image computing and computer-assisted intervention}, pages 234--241. Springer, 2015.

\bibitem[Shen et~al.(2023)Shen, Yang, Yu, Wong, Kaelbling, and Isola]{shen2023distilledfeaturefieldsenable}
William Shen, Ge~Yang, Alan Yu, Jansen Wong, Leslie~Pack Kaelbling, and Phillip Isola.
\newblock Distilled feature fields enable few-shot language-guided manipulation, 2023.
\newblock URL \url{https://arxiv.org/abs/2308.07931}.

\bibitem[Team et~al.(2023)Team, Anil, Borgeaud, Alayrac, Yu, Soricut, Schalkwyk, Dai, Hauth, Millican, et~al.]{team2023gemini}
Gemini Team, Rohan Anil, Sebastian Borgeaud, Jean-Baptiste Alayrac, Jiahui Yu, Radu Soricut, Johan Schalkwyk, Andrew~M Dai, Anja Hauth, Katie Millican, et~al.
\newblock Gemini: a family of highly capable multimodal models.
\newblock \emph{arXiv preprint arXiv:2312.11805}, 2023.

\bibitem[Tong et~al.(2022)Tong, Song, Wang, and Wang]{tong2022videomae}
Zhan Tong, Yibing Song, Jue Wang, and Limin Wang.
\newblock Videomae: Masked autoencoders are data-efficient learners for self-supervised video pre-training.
\newblock \emph{Advances in neural information processing systems}, 35:\penalty0 10078--10093, 2022.

\bibitem[Wang et~al.(2025)Wang, Chen, Huang, Dou, Liu, Gu, and Liu]{wang2025physctrl}
Chen Wang, Chuhao Chen, Yiming Huang, Zhiyang Dou, Yuan Liu, Jiatao Gu, and Lingjie Liu.
\newblock Physctrl: Generative physics for controllable and physics-grounded video generation.
\newblock In \emph{arXiv preprint}, 2025.

\bibitem[Wang et~al.(2020)Wang, Wei, Dong, Bao, Yang, and Zhou]{wang2020minilm}
Wenhui Wang, Furu Wei, Li~Dong, Hangbo Bao, Nan Yang, and Ming Zhou.
\newblock Minilm: Deep self-attention distillation for task-agnostic compression of pre-trained transformers.
\newblock \emph{Advances in neural information processing systems}, 33:\penalty0 5776--5788, 2020.

\bibitem[Xia et~al.(2024)Xia, Lin, Ma, and Wang]{xia2024video2game}
Hongchi Xia, Zhi-Hao Lin, Wei-Chiu Ma, and Shenlong Wang.
\newblock Video2game: Real-time, interactive, realistic and browser-compatible environment from a single video, 2024.

\bibitem[Xie et~al.(2023)Xie, Zong, Qiu, Li, Feng, Yang, and Jiang]{xie2023physgaussian}
Tianyi Xie, Zeshun Zong, Yuxing Qiu, Xuan Li, Yutao Feng, Yin Yang, and Chenfanfu Jiang.
\newblock Physgaussian: Physics-integrated 3d gaussians for generative dynamics.
\newblock \emph{arXiv preprint arXiv:2311.12198}, 2023.

\bibitem[Zhai et~al.(2024)Zhai, Shen, Chen, Wang, Wang, Wang, Guan, and Wang]{zhai2024physical}
Albert~J Zhai, Yuan Shen, Emily~Y Chen, Gloria~X Wang, Xinlei Wang, Sheng Wang, Kaiyu Guan, and Shenlong Wang.
\newblock Physical property understanding from language-embedded feature fields.
\newblock In \emph{Proceedings of the IEEE/CVF Conference on Computer Vision and Pattern Recognition}, pages 28296--28305, 2024.

\bibitem[Zhang et~al.(2025)Zhang, Li, Hauser, and Li]{zhang2025particle}
Kaifeng Zhang, Baoyu Li, Kris Hauser, and Yunzhu Li.
\newblock Particle-grid neural dynamics for learning deformable object models from rgb-d videos.
\newblock \emph{arXiv preprint arXiv:2506.15680}, 2025.

\bibitem[Zhang et~al.(2024)Zhang, Yu, Wu, Feng, Zheng, Snavely, Wu, and Freeman]{zhang2024physdreamer}
Tianyuan Zhang, Hong-Xing Yu, Rundi Wu, Brandon~Y. Feng, Changxi Zheng, Noah Snavely, Jiajun Wu, and William~T. Freeman.
\newblock {PhysDreamer}: Physics-based interaction with 3d objects via video generation.
\newblock In \emph{European Conference on Computer Vision}. Springer, 2024.

\bibitem[Zhong et~al.(2024)Zhong, Yu, Wu, and Li]{zhong2024springgaus}
Licheng Zhong, Hong-Xing Yu, Jiajun Wu, and Yunzhu Li.
\newblock Reconstruction and simulation of elastic objects with spring-mass 3d gaussians.
\newblock \emph{European Conference on Computer Vision (ECCV)}, 2024.

\end{thebibliography}
